%% file: 00_detok.tex
\documentclass[11pt]{article}

\usepackage{acl}

\usepackage{times}
\usepackage{latexsym}

\usepackage[T1]{fontenc}

\usepackage[utf8]{inputenc}

\usepackage{microtype}

\usepackage{inconsolata}

\usepackage{graphicx}

\usepackage{menukeys}

\usepackage{dirtytalk} 

\usepackage{xcolor}

\usepackage{amsmath}
\usepackage{booktabs}
\usepackage{placeins}
\usepackage{tikz}
\usetikzlibrary{trees,shadows,positioning,arrows.meta,shapes.misc,fit,calc, decorations.pathreplacing}
\usetikzlibrary{backgrounds}
\pgfdeclarelayer{background}
\pgfsetlayers{background,main}
\usepackage{xcolor}
\usepackage{pifont}     
\usepackage{amssymb}    

\newcommand{\Detok}{Detokenization}
\newcommand{\detok}{detokenization}
\newcommand{\sft}{FST}
\newcommand{\slt}{LST}
\newcommand{\Metric}{Canonicity}
\newcommand{\metric}{canonicity}
\newcommand{\metricsrc}{canonicity$_{\text{src}}$}
\usepackage{tcolorbox} 
\usetikzlibrary{matrix, decorations.pathreplacing}
\newtcolorbox{examplebox}[1][]{%
  colback=white,
  colframe=black,
  width=0.9\linewidth,
  boxrule=0.6pt,
  arc=2pt,
  auto outer arc,
  title=\textbf{Example},
  #1
}
\renewmenumacro{\keys}[,]{roundedkeys}
\changemenuelement{roundedkeys}{sep}{\hspace*{0.1em}}
\tikzset{
  roundedkeys/.append style={font=\ttfamily}
}
\newcommand{\tokens}[1]{\keys{#1}}

\title{Inside the LLM Word Factory}

\author{Benzi Busigin~~~~~~~Yuval Pinter \\
  Stein Faculty of Computer and Information Science \\
  Ben-Gurion University of the Negev \\
  Beer Sheva, Israel \\
  \texttt{\{busigin@post,uvp@cs\}.bgu.ac.il}}

\begin{document}
\maketitle
\begin{abstract}
Transformer language models process input provided as subword fragments, but natural language semantics usually rely on word-level concepts.
\emph{\Detok{}} is the process where models reconcile these two facts, aggregating subwords into word-level representations through their computation.
Prior work has found that this takes place mostly in early-to-middle layers, but so far the exact mechanics of the process have not been pinned down.
We venture deep into \detok{} using activation patching in controlled paired experiments that isolate the contribution of different model components, localizing English \detok{} in Llama2-7B to a two-stage process at Layer~1. 
Attention transmits a token-specific signal from nonfinal subwords, using sequential relays if necessary, while the MLP composes it with the local embedding.
This two-stage structure generalizes to twelve models from eight families, but the depth over which it takes place depends on the flavor of positional encoding: RoPE-based models detokenize over 1 to 5 layers, while learned-absolute models take 5 to 10.
Finally, we provide a probe for determining the success of the \detok{} process based on early-layer activations alone, performing at 0.94--0.97 AUROC depending on the amount of context.

\end{abstract}

\input{01_introduction}

\input{02_methodology}

\input{03_mechanism}

\input{04_scaling}

\input{05_generalization}

\input{06_predictability}

\input{07_conclusion}


\input{08_limitations}

\section*{Acknowledgments}
We thank Sarah Wiegreffe and Jacob Eisenstein, as well as the anonymous reviewers, for valuable comments.
This research was supported by the Israel Science Foundation (grant No. 1166/23) and by the Frenkel Foundation and the Stein Faculty of Computer Science and Information Systems.

\bibliography{custom,latex/anthology-1}

\clearpage
\appendix
\input{09_appendix}

\end{document}

%% file: 01_introduction.tex
\section{Introduction}

Subword tokenization is a near-universal preprocessing step in modern transformer language models~\citep{sennrich-etal-2016-neural,kudo-2018-subword,schmidt-etal-2024-tokenization}, yet it creates a systematic mismatch between input representation and internal computation: tokenizers split text into subword units based on statistical criteria, while the model's computations must operate over unified semantic concepts.
A given word can be tokenized in multiple ways; the word \say{table}, for instance, may be represented by the single token \tokens{\_table} or split into the pair \tokens{\_ta,ble}, and in the latter case the model must internally reconstruct a single representation of the concept \emph{table} from its fragments.
This internal reconstruction, termed \emph{\detok{}}~\citep{elhage2022, gurnee2023}, is a prerequisite for coherent downstream processing: when it fails, the model can lose access to the concept entirely.
\Detok{} is therefore a core internal computation that transformer language models must execute correctly on nearly every input.
Tokenization artifacts have been shown to produce behavioral failures ranging from anomalous outputs on under-trained tokens~\citep{land-bartolo-2024-fishing} to degraded performance on arithmetic and numerical reasoning~\citep{singh2024tokenization}.
These symptoms differ, but each one shows that tokenization can influence model behavior.
This motivates us to look beyond behavioral effects and examine the internal computation: how do transformer language models compose subword fragments back into the word-level concepts they encode, and what determines whether this composition succeeds or fails for a given input?

\input{plots/patching_diagram}

Recent work has made substantial progress on understanding \detok{}.
Under the \emph{stages-of-inference} framing, early layers of transformer language models are thought to detokenize subword input into an internal vocabulary on which later stages operate~\citep{lad2024stages}.
\citet{kaplan2025tokens} provide the most comprehensive account to date, showing that word-level representations become decodable from the last subword position in early-to-middle layers, and describing a broad two-stage process: attention aggregates subword tokens, and feed-forward layers (MLP) retrieve a word-level representation.
Complementary evidence comes from \citet{feucht-etal-2024-token}, who show that last-token positions of multi-token words \say{erase} information about preceding tokens in early layers.
\citet{kamoda-etal-2025-weight} use a different method: an input-agnostic, weight-based analysis of first-layer attention in GPT-2, which finds structure supporting \detok{}.
These studies establish that \detok{} is a core process in transformer language models to which both attention and MLPs contribute.
This understanding is already actionable: \citet{mehta2026defragmenting} use \detok{} to guide vocabulary expansion for under-served scripts, deciding both which items to add and how to initialize their embeddings.

Existing evidence characterizes \detok{} at the level of entire layers or broad component types, relying predominantly on correlational probing or coarse-grained interventions.
Three questions remain open.
First, within the critical early layers, which \textbf{specific} sub-components drive the process, and what signal do they transmit?
Second, does the mechanism \textbf{generalize} across architectures and word lengths, and what determines how many layers it spans?
Third, can the outcome of \detok{}, i.e., whether it will succeed or fail for a given input, be \textbf{predicted} from the early-layer representation, and would such a prediction hold for natural text as well as for controlled inputs?
We address these questions through activation patching combined with a controlled paired-dataset design that, by construction, isolates the causal contribution of attention from that of the MLP.

On English data, across twelve transformer language models, we find that Layer~1 implements a two-stage computation that initiates \detok{}.
First, a small subset of attention heads transmit a weak, input-dependent signal from the preceding token into the last subword position.
Then, Layer~1's MLP composes this signal with the local subword embedding through a continuous transformation.
The same division of labor holds across word lengths from two to six subword tokens, with composition still occurring at the last position and intermediate positions serving as relays.
\Detok{} concentrates in the first 6--30\% of layers across all architectures, with the depth governed by the positional encoding strategy.
The early-layer signal is linearly readable: a single direction fit at the depth identified by activation patching predicts final-layer \detok{} success with AUROC 0.94 on isolated, contextless sequences, and 0.97 on natural text.
Together, these results identify a specific, generalizable mechanism for \detok{}, whose depth is governed by positional encoding rather than scale, and provide an early-layer linear predictor of \detok{} success that transfers to natural text.\footnote{We release our code and data at
\url{https://github.com/MeLeLBGU/Word-Factory}.}

%% file: plots/patching_diagram.tex
\begin{figure*}[t]
\centering

\definecolor{sharedC}{RGB}{31,119,180}
\definecolor{variesA}{RGB}{214,96,40}
\definecolor{variesB}{RGB}{30,140,120}
\definecolor{inertC}{RGB}{120,120,120}
\definecolor{successC}{RGB}{46,125,50}
\definecolor{failC}{RGB}{183,28,28}
\definecolor{patchC}{RGB}{120,40,140}

\begin{tikzpicture}[
  font=\small,
  layer/.style={draw, rounded corners=1.5pt, minimum width=8mm, minimum height=5.5mm,
                fill=inertC!10, draw=inertC!60, inner sep=1.5pt, font=\scriptsize},
  attnNode/.style={draw, rounded corners=1.5pt, minimum width=10mm, minimum height=5.5mm,
                   fill=inertC!10, draw=inertC!60, inner sep=1.5pt, font=\scriptsize},
  mlpNode/.style={draw, rounded corners=1.5pt, minimum width=10mm, minimum height=5.5mm,
                  fill=inertC!10, draw=inertC!60, inner sep=1.5pt, font=\scriptsize},
  tok/.style={draw, rounded corners=1pt, minimum width=8mm, minimum height=5mm,
              font=\scriptsize\ttfamily, inner sep=1.5pt},
  bostok/.style={tok, fill=inertC!8, draw=inertC!50, font=\tiny\ttfamily,
                 minimum width=6mm},
  tokShared/.style={tok, fill=sharedC!20, draw=sharedC},
  tokVarA/.style={tok, fill=variesA!22, draw=variesA},
  tokVarB/.style={tok, fill=variesB!22, draw=variesB},
  tokPlain/.style={tok, fill=white, draw=inertC!70},
  resid/.style={draw, circle, minimum size=4mm, inner sep=0pt,
                fill=inertC!10, draw=inertC!70, font=\tiny},
  residA/.style={resid, fill=variesA!28, draw=variesA, very thick},
  residB/.style={resid, fill=variesB!28, draw=variesB, very thick},
  residS/.style={resid, fill=sharedC!28, draw=sharedC, very thick},
  flow/.style={->, >=Stealth, thin, inertC!70},
  meas/.style={<->, >=Stealth, semithick, successC, dashed},
  patch/.style={->, >=Stealth, very thick, patchC},
  panelTitle/.style={font=\small\bfseries, anchor=west},
  pairLbl/.style={font=\scriptsize\bfseries, sharedC!50!black, anchor=west},
  frame/.style={
    draw=inertC!40,
    fill=white,
    rounded corners=6pt,
    line width=0.5pt,
    inner sep=4pt
    },
]

\begin{scope}[local bounding box=panelA]
  \node[panelTitle] at (-0.2, 2) {(a) Setup};

  \begin{scope}[yshift=0.2cm] 
    \node[font=\scriptsize\itshape, inertC!80!black, anchor=west] at (0, 1.4) {canonical};
    \node[tokPlain, anchor=west] (Acan) at (0, 0.95) {\_folder};
    \node[layer, right=1.5mm of Acan] (AcL0) {L0};
    \node[layer, right=0.8mm of AcL0] (AcD)  {$\cdots$};
    \node[layer, right=0.8mm of AcD]  (AcL30){L30};
    \draw[flow] (Acan) -- (AcL0);
    \draw[flow] (AcL0) -- (AcD);
    \draw[flow] (AcD)  -- (AcL30);
    \node[tokPlain, draw=inertC!80, thick, anchor=west] (Acan) at (0, 0.95) {\_folder};

    \node[font=\scriptsize\itshape, inertC!80!black, anchor=west] at (0, 0.4) {split};
    \node[tokPlain, fill=white, anchor=west] (Asp1) at (0, -0.05) {\_fo};
    \node[tokPlain, draw=inertC!80, thick, right=0.4mm of Asp1] (Asp2) {lder};
    \node[font=\tiny\itshape, inertC, anchor=north, yshift=-0.1mm]
      at (Asp2.south) {last subword pos.};
    \node[layer, right=1.5mm of Asp2] (AsL0) {L0};
    \node[layer, right=0.8mm of AsL0] (AsD)  {$\cdots$};
    \node[layer, right=0.8mm of AsD]  (AsL30){L30};
    \draw[flow] (Asp2.east) -- (AsL0.west);
    \draw[flow] (AsL0) -- (AsD);
    \draw[flow] (AsD)  -- (AsL30);

    \draw[meas] (AcL30.south) -- node[pos=0.1, right=2.2mm, font=\scriptsize, successC, align=center]
      {\textbf{\Metric} \\ cos\_sim (\_folder, lder)} (AsL30.north);
  \end{scope}

  \path (-0.3, 0) -- (6.5, 0);
\end{scope}

\begin{scope}[shift={(0, -2.5)}, local bounding box=panelB]
  \node[panelTitle] at (-0.2, 1.4) {(b) Pair examples};
  \node[pairLbl] at (0.0, 0.9) {\textsc{LST}};
  \node[tokPlain] (BSLs1) at (0.4, 0.4) {\_po};
  \node[tokShared, right=0.4mm of BSLs1] (BSLs2) {or};
  \node[font=\scriptsize, right=1.2mm of BSLs2, successC] {0.91\,\checkmark};
  \node[tokPlain] (BSLf1) at (0.4, -0.25) {\_err};
  \node[tokShared, right=0.4mm of BSLf1] (BSLf2) {or};
  \node[font=\scriptsize, right=1.2mm of BSLf2, failC] {0.48\,\ding{55}};
  \node[font=\tiny, inertC, below=0.2mm of BSLf1] {pos 1};
  \node[font=\tiny, inertC, below=0.2mm of BSLf2] {pos 2};
  \node[pairLbl] at (3.2, 0.9) {\textsc{FST}};
  \node[tokShared] (BSFs1) at (3.6, 0.4) {\_fo};
  \node[tokPlain, right=0.4mm of BSFs1] (BSFs2) {lder};
  \node[font=\scriptsize, right=1.2mm of BSFs2, successC] {0.94\,\checkmark};
  \node[tokShared] (BSFf1) at (3.6, -0.25) {\_fo};
  \node[tokPlain, right=0.4mm of BSFf1] (BSFf2) {od};
  \node[font=\scriptsize, right=1.2mm of BSFf2, failC] {0.51\,\ding{55}};
  \node[font=\tiny, inertC, below=0.2mm of BSFf1] {pos 1};
  \node[font=\tiny, inertC, below=0.2mm of BSFf2] {pos 2};
\path (-0.3, 0) -- (6.5, 0);
\end{scope}

\begin{scope}[shift={(7.95, -3.38)}, local bounding box=panelC,
              x=0.80cm, y=0.62cm,
              every node/.append style={transform shape=false}]

  \tikzset{
    cellSrc/.style    ={draw, rounded corners=1pt, minimum width=5.5mm, minimum height=4mm,
                        fill=variesA!18, draw=variesA!70, line width=0.4pt, inner sep=0pt},
    cellSrcEmb/.style ={draw, rounded corners=1pt, minimum width=5.5mm, minimum height=4mm,
                        fill=inertC!8,   draw=inertC!60, line width=0.4pt, inner sep=0pt},
    cellTgt/.style    ={draw, rounded corners=1pt, minimum width=5.5mm, minimum height=4mm,
                        fill=variesB!18, draw=variesB!70, line width=0.4pt, inner sep=0pt},
    cellTgtEmb/.style ={draw, rounded corners=1pt, minimum width=5.5mm, minimum height=4mm,
                        fill=inertC!8,   draw=inertC!60, line width=0.4pt, inner sep=0pt},
    cellPatched/.style={draw, rounded corners=1pt, minimum width=5.5mm, minimum height=4mm,
                        fill=patchC!18,  draw=patchC,    line width=0.4pt, inner sep=0pt},
    cellDonor/.style  ={draw, rounded corners=1pt, minimum width=5.5mm, minimum height=4mm,
                        fill=variesA!35, draw=variesA,   line width=1.2pt, inner sep=0pt},
    cellRecip/.style  ={draw, rounded corners=1pt, minimum width=5.5mm, minimum height=4mm,
                        fill=patchC!30,  draw=patchC,    line width=1.2pt, dashed, inner sep=0pt},
    gridTok/.style    ={font=\tiny\ttfamily, text=inertC!90!black, inner sep=1pt},
    gridLyr/.style    ={font=\tiny, text=inertC!90!black, anchor=east, inner sep=1pt},
    gridTtl/.style    ={font=\scriptsize\bfseries, inner sep=1pt},
    residArr/.style   ={->, >=Stealth, draw=inertC!35, line width=0.25pt},
    patchSite/.style    ={draw, circle, minimum size=1.4mm, inner sep=0pt,
                          line width=0.3pt},
    patchSiteSrc/.style ={patchSite, fill=variesA!75, draw=variesA},
    patchSiteTgt/.style ={patchSite, fill=variesB!75, draw=variesB},
    patchSiteHit/.style ={draw, rectangle, minimum size=1.6mm, inner sep=0pt,
                          fill=patchC, draw=patchC!50!black, line width=0.5pt},
    siteLbl/.style      ={font=\tiny\ttfamily, text=inertC!90!black, inner sep=1pt},
    patchSiteCont/.style ={patchSite, fill=patchC!60, draw=patchC},
  }

  \def\yEmbed{1}
  \def\yLzero{2}
  \def\yLone{3.5}
  \def\yDots{5}
  \def\yLnm2{6}
  \def\srcShift{-0.5}

  \def\yResid{3.85}
  \def\yMlp{3.50}
  \def\yAttn{3.15}

  \node[panelTitle, anchor=west] at (-0.6, 8.6)
    {(c) Patching L1 \texttt{attn\_out} at pos.~2 (\textsc{LST} pair)};

  \foreach \c in {1,...,3}{
    \node[cellSrcEmb] (Sc1-\c) at (\c+\srcShift, \yEmbed) {};
  }
  \foreach \c in {1,...,3}{
    \node[cellSrc] (Sc2-\c) at (\c+\srcShift, \yLzero) {};
  }
  \foreach \c in {1,2}{
    \node[cellSrc, minimum height=11mm] (Sc3-\c) at (\c+\srcShift, \yLone) {};
  }
  \node[cellDonor, minimum height=11mm] (Sc3-3) at (3+\srcShift, \yLone) {};
  \foreach \c in {1,...,3}{
    \node[cellSrc] (Sc4-\c) at (\c+\srcShift, \yDots) {};
  }
  \foreach \c in {1,...,3}{
    \node[cellSrc] (Sc5-\c) at (\c+\srcShift, \yLnm2) {};
  }

  \foreach \c in {1,2}{
    \node[patchSiteSrc] at (\c+\srcShift, \yResid) {};
    \node[patchSiteSrc] at (\c+\srcShift, \yMlp)   {};
    \node[patchSiteSrc] at (\c+\srcShift, \yAttn)  {};
  }
  \node[patchSiteSrc] at (3+\srcShift, \yResid) {};
  \node[patchSiteSrc] at (3+\srcShift, \yMlp)   {};
  \node[patchSiteHit] (SaO3) at (3+\srcShift, \yAttn) {};

  \node[siteLbl, anchor=west, font=\tiny] at (3.45+\srcShift, \yResid) {resid\_post};
  \node[siteLbl, anchor=west, font=\tiny] at (3.45+\srcShift, \yMlp)   {mlp\_out};
  \node[siteLbl, anchor=west, font=\tiny] at (3.45+\srcShift, \yAttn)  {attn\_out};

  \node[gridTok] at (1+\srcShift, 0.30) {[BOS]};
  \node[gridTok] at (2+\srcShift, 0.30) {\_po};
  \node[gridTok] at (3+\srcShift, 0.30) {or};

  \node[gridTtl, variesA] at (2+\srcShift, 7.0) {Source run};

  \foreach \c in {5,...,7}{
    \node[cellTgtEmb] (Tc1-\c) at (\c, \yEmbed) {};
  }
  \foreach \c in {5,...,7}{
    \node[cellTgt] (Tc2-\c) at (\c, \yLzero) {};
  }
  \foreach \c in {5,6}{
    \node[cellTgt, minimum height=11mm] (Tc3-\c) at (\c, \yLone) {};
  }
  \node[cellRecip, minimum height=11mm] (Tc3-7) at (7, \yLone) {};
  \foreach \c in {5,6}{
    \node[cellTgt] (Tc4-\c) at (\c, \yDots) {};
  }
  \node[cellPatched] (Tc4-7) at (7, \yDots) {};
  \foreach \c in {5,6}{
    \node[cellTgt] (Tc5-\c) at (\c, \yLnm2) {};
  }
  \node[cellPatched] (Tc5-7) at (7, \yLnm2) {};

  \foreach \c in {5,6}{
    \node[patchSiteTgt] at (\c, \yResid) {};
    \node[patchSiteTgt] at (\c, \yMlp)   {};
    \node[patchSiteTgt] at (\c, \yAttn)  {};
  }
  \node[patchSiteCont] at (7, \yResid) {};
  \node[patchSiteCont] at (7, \yMlp)   {};
  \node[patchSiteHit] (TaO7) at (7, \yAttn) {};

  \node[gridTok] at (5, 0.30) {[BOS]};
  \node[gridTok] at (6, 0.30) {\_err};
  \node[gridTok] at (7, 0.30) {or};

  \node[gridTtl, variesB] at (6, 7.0) {Target run};

  \node[gridLyr] at (0.55+\srcShift, \yEmbed) {embed};
  \node[gridLyr] at (0.55+\srcShift, \yLzero) {layer 0};
  \node[gridLyr] at (0.55+\srcShift, \yLone)  {layer 1};
  \node[gridLyr] at (0.55+\srcShift, \yDots + 0.2)  {$\vdots$};
  \node[gridLyr] at (0.55+\srcShift, \yLnm2)  {layer $n{-}2$};

  \foreach \c in {1,...,3}{
    \draw[residArr] (Sc1-\c.north) -- (Sc2-\c.south);
    \draw[residArr] (Sc2-\c.north) -- (Sc3-\c.south);
    \draw[residArr] (Sc3-\c.north) -- (Sc4-\c.south);
    \draw[residArr] (Sc4-\c.north) -- (Sc5-\c.south);
  }
  \foreach \c in {5,...,7}{
    \draw[residArr] (Tc1-\c.north) -- (Tc2-\c.south);
    \draw[residArr] (Tc2-\c.north) -- (Tc3-\c.south);
    \draw[residArr] (Tc3-\c.north) -- (Tc4-\c.south);
    \draw[residArr] (Tc4-\c.north) -- (Tc5-\c.south);
  }

  \draw[patch, rounded corners=4pt, line width=0.9pt]
    (SaO3.east) -- ++(0.30,0)
    -- ($(SaO3.east)+(0.30,-0.4)$)
    -- ($(TaO7.east)+(0.30,-0.4)$)
    -- ($(TaO7.east)+(0.30,0)$)
    -- (TaO7.east);

  \coordinate (CosSrcTip) at (3+\srcShift, 7.55);
  \draw[meas, ->] (Sc5-3.north) -- ++(0,0.25) -- (CosSrcTip);
  \node[font=\scriptsize, successC, anchor=south, inner sep=0.5pt] at (CosSrcTip)
    {\metric(or, \_poor)};

  \coordinate (CosTgtTip) at (7, 7.55);
  \draw[meas, ->] (Tc5-7.north) -- ++(0,0.25) -- (CosTgtTip);
  \node[font=\scriptsize, successC, anchor=south, inner sep=0.5pt] at (CosTgtTip)
    {\metricsrc(or, \_poor)};
\end{scope}

\begin{scope}[on background layer]
  \node[frame, inner sep=4pt, fit=(panelA)] {};
  \node[frame, inner ysep=2pt, fit=(panelB)] {};
  \node[frame, inner sep=6pt, fit=(panelC)] {};
\end{scope}

\begin{scope}[shift={(0, -4)}]
  \node[tokShared, minimum width=5mm, minimum height=4mm] (Lsh) at (0, 0) {};
  \node[anchor=west, font=\scriptsize] (Lsh_t) at (Lsh.east) {\,shared};
  \node[tokVarA, minimum width=5mm, minimum height=4mm, right=4mm of Lsh_t.east] (LvA) {};
  \node[anchor=west, font=\scriptsize] (LvA_t) at (LvA.east) {\,source identity};
  \node[tokVarB, minimum width=5mm, minimum height=4mm, right=4mm of LvA_t.east] (LvB) {};
  \node[anchor=west, font=\scriptsize] (LvB_t) at (LvB.east) {\,target identity};
  \coordinate (Lp_start) at ([xshift=4mm]LvB_t.east);
  \draw[patch] (Lp_start) -- ++(5mm, 0) coordinate (Lp_end);
  \node[anchor=west, font=\scriptsize] at (Lp_end) {\,patch};
\end{scope}

\end{tikzpicture}

\caption{
\textbf{Experimental design and activation patching.}
\textbf{(a)}~Each word the tokenizer renders as a single token is run twice: as the canonical single token and as an artificial split. \emph{\Metric{}} compares the layer-30 residual streams at the canonical position and the split's last subword position.
\textbf{(b)}~\slt{} pairs share the second token; \sft{} pairs share the first. Each pair has one successful and one failed split, with \metric{} values shown.
\textbf{(c)}~Activation patching at L1 \texttt{attn\_out}, position~2, for an \slt{} pair. The source run \mbox{[\tokens{BOS}, \tokens{\_po,or}]} reconstructs \tokens{\_poor}. Replacing the target's L1 \texttt{attn\_out} at position~2---which normally aggregates the target context \tokens{\_err,or}---with the source's, then completing the forward pass unchanged, makes the target's L30 residual align with \tokens{\_poor} (the \emph{source's} word) rather than \tokens{\_error}. 
A single L1 attention vector thus carries enough cross-position signal to make the target's run reconstruct the source's word instead of its own.
}
\label{fig:design}
\end{figure*}
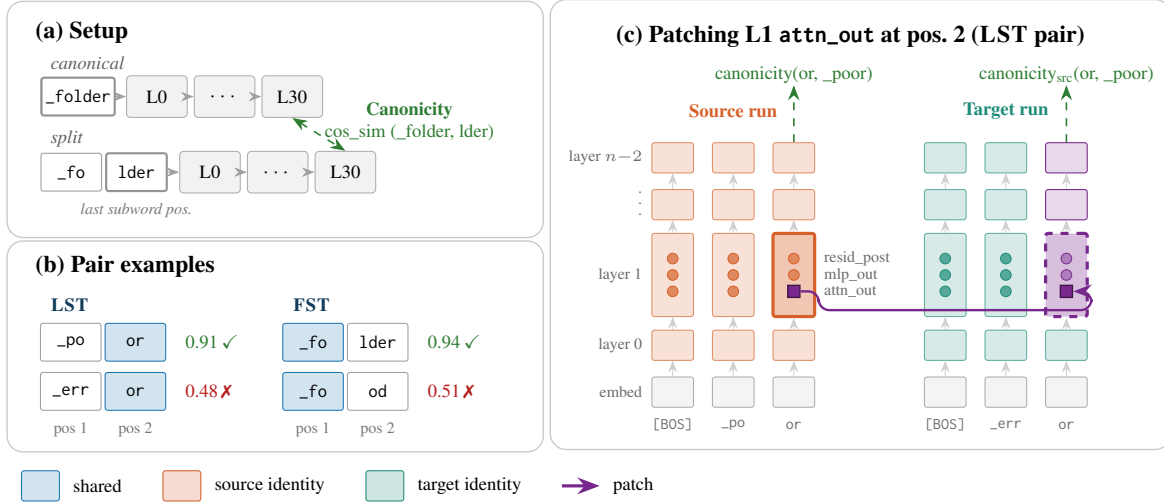

%% file: 02_methodology.tex
\section{Setup}
\label{sec:methodology}
 
Many words that a tokenizer represents as a single token can also be split into multiple subword tokens.
For example, the word \say{table} may be tokenized as a single token \tokens{\_table} or split into \tokens{\_ta,ble}.
When a word is split, the model must internally reconstruct the full-word meaning from its parts, a process known as \emph{\detok{}}.
We study this process by leveraging words that admit both representations: a canonical single-token form and an artificially-split multi-token form.
This dual representation provides a ground truth: we can compare what the model computes from the split input against the representation it produces for the intact word.
Our analysis is therefore restricted to words that the tokenizer can represent as a single token; inherently multi-token words lack such a reference and are excluded.
All experiments before \autoref{sec:scaling} consider only two-token splits.

\paragraph{Data.}
For each model, we collect a pool of \emph{single-token words}: English words that exist in the model's tokenizer vocabulary as one token.
Restricting to such words is necessary because our analysis compares a split tokenization of a word against its canonical single-token representation; inherently multi-token words allow no such reference.
We draw candidate words from three publicly available English word lists: a frequency-ordered list of the 100{,}000 most common lowercased English words from the \texttt{top-english-wordlists} repository; 
the Google 10{,}000-word common English list; 
and the English word frequency dataset~\citep{tatman2017englishfreq}, a unigram frequency table of approximately 333{,}000 entries derived from the Google Web Trillion Word Corpus. 
More information, as well as details about preprocessing, is available in \autoref{app:data_sources}.

\paragraph{Notation.} We adopt standard transformer-component notation throughout.
We refer to the attention block output as \texttt{attn\_out} and the MLP block output as \texttt{mlp\_out}.
\texttt{resid\_pre} denotes the residual stream at the input of a layer, and \texttt{resid\_post} the residual stream at its output, after both the attention and the MLP blocks have written to it.
We index layers from zero, therefore
\texttt{resid\_pre} at Layer~0 is the token embedding alone, while \texttt{resid\_post} at Layer~0 already contains the output of one complete transformer block.
Unless otherwise stated, all activations are taken at the last subword position of the split word.

\paragraph{Model.}
All primary experiments use Llama~2--7B~\citep{touvron2023llama2} with 32~layers, $d_{\text{model}} = 4{,}096$, with a word pool of size 8,933.
We discuss cross-architecture generalities in \autoref{sec:cross_model}.

\subsection{Measuring Success of \Detok{}}
\label{sec:metric}
We introduce our primary success metric, \emph{\metric{}}, defined as the cosine similarity, at layer $n-2$ (layer~30 of~32 in Llama2-7B), between \texttt{resid\_post} at the last subword position of the split input and the corresponding canonical single-token representation at the same layer and position (see \autoref{fig:design}a).
High \metric{} indicates that the model has successfully reconstructed the full-word representation from fragments; a low value indicates failure.
We read at layer $n-2$ because it is late enough for the computation to have largely stabilized, and early enough to precede the final layer, which reshapes the residual stream for next-token prediction~\citep{belrose2023eliciting}.
A per-layer sweep supports the choice: the correlation between representational and behavioral similarity plateaus across the late layers in every model tested, reaching $\rho_{\text{KL}} = -0.86$ at layer $n-2$ on Llama2-7B, so $n-2$ represents the plateau instead of a tuned optimum.
This choice of layer fixes \emph{where} we read. 
It remains to show that \emph{what} we read there tracks behavior. 
\Metric{} is a strong, monotonic predictor of model behavior rather than a geometric artifact. 
Across our full word pool, split inputs in the top \metric{} quintile match the canonical input's top-1 next-token prediction 78\% of the time, versus 24\% in the bottom quintile.

We demonstrate the cost of low \metric{} using two splits of the same word.
Both \tokens{\_ad,vanced} and \tokens{\_advance,d} spell \say{advanced}, but only the former matches the canonical next-token distribution (KL 0.05) with predicted continuations such as \say{technology} and \say{manufact}.
The latter predicts \say{raft} and \say{iet}, suggesting that the model reads the final subword \tokens{d} as the beginning of a new word.
We provide both correlational and causal validation of \metric{}, together with further qualitative examples, in \autoref{app:metric_validation}.

\subsection{Experimental Design}
\label{sec:datasets}

The success of \detok{} depends on both subword tokens. 
To isolate the contribution of each component, we construct controlled pairs of split words that share one token and differ on the other.

\begin{figure*}
    \centering
    \includegraphics[width=1\linewidth]{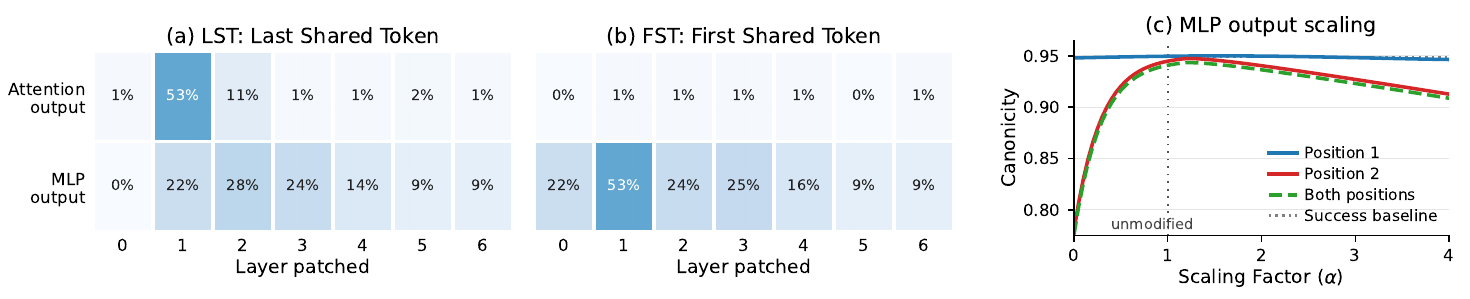}
    \caption{Activation patching at the second token. Values show the percentage of the \metricsrc{} gap closed. \textbf{(a)} In \slt{}, Layer~1 \texttt{attn\_out} drives the effect.
  \textbf{(b)} In \sft{}, attention is uninformative, and Layer~1 \texttt{mlp\_out} dominates.
  \textbf{(c)} Scaling the Layer~1 \texttt{mlp\_out} by a factor $\alpha$ reveals a smooth, continuous effect on \detok{} quality. Scaling position~2 raises \metric{} from 0.78 ($\alpha=0$) to the source baseline ($\alpha=1$); scaling position~1 has little effect.}

  \label{fig:mechanism_overview}
\end{figure*}

\paragraph{Pair construction.}
Each pair consists of two split words that share exactly one token position.
Within each pair, one member has high \metric{} in our chosen model and one has low \metric{}; we refer to these as the \emph{successful} and \emph{failed} members of the pair, respectively, and denote the difference in their \metric{} as the \emph{success--failure gap}.
We define two complementary settings (\autoref{fig:design}b):

\begin{itemize}
    \item \textbf{First-Shared-Token (\sft{}):} pairs share the first token and differ in the second.
    For example, the split \tokens{\_fo,lder} reconstructs the representation of \tokens{\_folder} well (\metric{} 0.94, \emph{successful}), while the split \tokens{\_fo,od} reconstructs the representation of \tokens{\_food} poorly (\metric{} 0.51, \emph{failed}).
    The pair members share the first token \tokens{\_fo}.
    \item \textbf{Last-Shared-Token (\slt{}):} pairs share the second token and differ in the first.
    For example, the split \tokens{\_po,or} reconstructs \tokens{\_poor} well (\metric{} of 0.91, \emph{successful}), while the split \tokens{\_err,or} reconstructs \tokens{\_error} poorly (\metric{} of 0.48, \emph{failed}).
    The pair members share the second token \tokens{or}.
\end{itemize}

The experimental value of this design lies in what each setting holds constant. 
In \sft{}, the first token, and therefore the value vector at position~1, is identical across the pair, so the cross-position contribution that any attention head can read from the first token is approximately matched between runs. 
Any difference in outcome must therefore arise from how subsequent components, including the MLP, process the differing local embedding at the last position. 
In \slt{}, the local embedding at the last position is identical across the pair, so the two members are matched in everything an MLP receives from the local position; any difference must arise from cross-position information that attention transmits from the differing first token.
This complementary structure separates the contributions of attention and MLP, and is the foundation of the analysis in \autoref{sec:mechanism}.

Pairs are constructed by requiring a minimum success--failure gap in \metric{} between the two members.
Because \metric{} spreads vary across architectures~\citep{ethayarajh-2019-contextual}, we calibrate this threshold per model to ensure a sufficient number of pairs for each dataset.
For Llama2-7B (gap $\geq 0.30$), this yields $N=6{,}409$ \slt{} and $N=6{,}254$ \sft{} pairs; per-model thresholds and counts are listed in \autoref{app:data_sources}.

\subsection{Activation Patching}
\label{sec:patching_method}
Throughout the paper, we use activation patching~\citep{vig2020causal, geiger2021causal, meng2022rome} to localize the components responsible for \detok{}.
Given a pair from the dataset, we patch a chosen component's activation from the \emph{source} (successful member) into the \emph{target} (failed member), complete the target's forward pass, and measure whether the source's canonical representation now emerges in the target (\autoref{fig:design}c).
A high score indicates that the patched component carries the composed concept, or information used in its composition.
We measure this as \emph{\metricsrc{}}: the layer-$n-2$ cosine similarity between the patched target and the \emph{source}'s canonical representation.
We report the fraction of the gap that the patch closes between the target baseline (\metricsrc{} of the unpatched target run) and the source baseline (the source's standard \metric{}).
0\% means the patch had no effect; 100\% means the patched target reaches the source baseline, indicating that the source concept has been fully reconstructed in the target.

%% file: 03_mechanism.tex
\section{The Two-Stage Mechanism}
\label{sec:mechanism}
 
We now turn to the core mechanistic question: how does the model compose two subword tokens into a single semantic representation?
All experiments in this section use Llama2-7B on two-token words from the \slt{} and \sft{} pair sets. 
The unpatched source and target baselines (\metricsrc{}) 
are approximately 0.95 and 0.43, respectively, in both pair sets.
We will refer to the tokens and their representations by their position, 1 and 2.

\subsection{Activation Patching Localizes the Mechanism to Layer~1}
\label{sec:patching}

To localize where information from position~1 first reaches position~2, we perform activation patching in the \slt{} setting.
For each layer independently, we patch a single component's activation at position~2 from source into target. 
\autoref{fig:mechanism_overview}a,b reports the fraction of the gap closed.
Prior to Layer~1, patching any component has no effect: position~2 contains only the local token embedding and has not yet received information from position~1.
Patching the Layer~1 attention output, however, closes 53\% of the gap, the largest single-component effect at any layer; later layers yield minimal improvement.
Layer~1 attention is thus the first and primary mechanism by which information from the first token reaches position~2.

This effect is specific to \detok{}, not a generic property of early-layer attention.
The same intervention, applied to ordinary two-word sequences that share a last word and require no composition (e.g., \tokens{the,cat} vs.\ \tokens{a,cat}), closes 6.2\% of the gap, against 50.8\% for split-word pairs under a metric adapted to both conditions (\autoref{app:detok_specificity}).
Layer~1 attention therefore transmits cross-position information in ordinary sequences too, but its causal contribution is about eight times larger during \detok{}.

The MLP's contribution is smaller and more distributed, peaking at Layer~2 (28\% gap closed) and declining gradually thereafter. 
No single MLP layer matches the L1 attention effect.
However, since the MLPs read from a residual stream that L1 attention has already written to, \slt{} alone cannot tell us whether the MLP plays a distinct role in the process or merely passes along the attention signal.
We turn to \sft{} to find out.

\subsection{MLP and Attention Have Distinct Roles}
\label{sec:\sft_asymmetry}
 
In \sft{}, source and target share the first token, so the value vectors at position~1 are identical by construction.
The Layer~1 attention output at position~2 is not: the differing second-token embedding can change the query and the attention pattern there.
We therefore measure it directly.
Across all 6{,}254 \sft{} pairs, the within-pair cosine similarity of L1 \texttt{attn\_out} at position~2 averages 0.95, against 0.74 for words with a different first token (\autoref{app:fst_attn_similarity}).
The intervention that drove most of the effect in \slt{} therefore has almost nothing to transfer, and indeed patching L1 \texttt{attn\_out} in \sft{} closes only 1\% of the gap.
The reverse intervention is sharper: injecting a target run's \texttt{attn\_out} into a source run barely degrades performance (\metricsrc{} 0.92 vs.\ 0.94).
The L1 attention output is genuinely interchangeable across the pair when the first token is the same; it carries no information that distinguishes successful from failed composition.
What does distinguish them is the MLP.
Patching L1 \texttt{mlp\_out} closes 53\% of the gap, more than it closed in \slt{}, and the only intervention that produces meaningful reconstruction in this setting.
Because the pre-MLP residual stream differs between the runs only via the second-token embedding, the MLP's nonlinear transformation is what turns this local input into a successful (or failed) reconstruction.
Together, these results establish a two-stage process: attention transmits cross-position information, and the MLP composes it with the local embedding.

\subsection{Attention as a Directional Nudge}
\label{sec:ov_circuit}
We established that L1 attention transmits cross-position information from the first token, but not what that information is.
We decompose the L1 attention output to find out.

\paragraph{The signal flows through a few heads, via values at position~1.}
Patching individual per-head, pre-projection output vectors $\mathbf{z}^{(h)}$ localizes the effect to three heads: head~27 alone closes 17\% of the gap, and heads~24 and~28 a further 7\% when patched together.
The three together close 39\% of the gap, indicating a super-additive interaction in which the heads act jointly rather than independently.
This recovers most of L1 \texttt{attn\_out}'s 53\% effect, with the remainder spread thinly across the other heads.
Component-wise, the effect is carried almost entirely by \emph{value} vectors at position~1; patching queries, keys, or attention patterns produces negligible change.

\paragraph{The signal points toward the canonical embedding, but only weakly.}
These three dominant heads write vectors weakly but consistently aligned with the canonical embedding direction (mean cosine similarity 0.053, vs.\ $-$0.01 for the other heads), and their combined projection moderately predicts \detok{} success (Spearman $\rho = 0.52$).
Layer~1 attention thus writes a small, input-dependent nudge toward the canonical embedding.

\paragraph{The signal is token-specific, not a generic success flag.}
A natural concern is that the L1 attention output might encode a generic successful composition flag.
The \sft{} results rule this out: there, source and target share the first token, so L1 attention writes nearly the same vector in both runs, yet they diverge in composition success.
Whatever attention writes is therefore tied to the first-token identity, not to the success of the eventual composition.
Together, these results refine the two-stage picture: L1 attention does not encode the composed concept.
Rather, it writes a small, token-specific signal that the MLP uses to assemble it.

\subsection{The MLP Composes Continuously}
\label{sec:mlp_composition}

Having characterized the attention signal, we turn to the next stage: how does the Layer~1 MLP use this signal to compose the full-word representation? 
Unlike the patching experiments, the interventions described below modify a single run instead of transferring activations between runs; we therefore report raw \metric{}, not gap-closed percentages.

\paragraph{Necessity.} 
Zero-ablating L1 \texttt{mlp\_out} at position~2 drops \metric{} from 0.95 to 0.78, the single largest MLP ablation effect at any layer; ablating L0 \texttt{mlp\_out} causes comparable damage.
Beyond Layer~3, zero-ablation has negligible impact (\metric{} > 0.93); the critical MLP computation is concentrated in the earliest layers. 

\paragraph{Continuous operation.} 
To test whether the MLP functions as a discrete gate or a continuous transformation, we scale L1 \texttt{mlp\_out} by a factor $\alpha \in [0, 4]$ (\autoref{fig:mechanism_overview}c). 
While scaling only position~1 has virtually no effect, scaling position~2, or both positions jointly, produces a smooth response: \metric{} rises steadily from 0.78 at $\alpha = 0$ to 0.95 at $\alpha = 1$, then declines. 
The absence of any threshold, plateau, or step, rules out a discrete key-value retrieval account of the per-input computation~\citep{geva-etal-2021-transformer,geva-etal-2022-transformer,kaplan2025tokens}.

\paragraph{}
Together with the attentional findings (\S\ref{sec:ov_circuit}), this gives a complete picture of the Layer~1 mechanism: attention writes a small, token-specific directional signal, and the MLP applies a continuous transformation that turns this signal plus the local embedding into the canonical representation.

%% file: 04_scaling.tex
\section{Scaling with Token Count}
\label{sec:scaling}

We have established that for two-token words, the two-stage mechanism of \detok{} executes primarily at Layer~1.
Words split into more tokens require more layers in practice; we ask why. 
We extend the controlled-pairs dataset to $k$-token splits ($k \in [2,6]$), focusing on the most demanding configuration, where the only differing token is the first subword,\footnote{An example for $k=4$ would be \tokens{\_ant,ic,ipa,tion} and \tokens{\_part,ic,ipa,tion}.} and examine how the number of required layers grows with token count, and what the mechanism behind that growth is.
These are still in-vocabulary words, split artificially into $k$ tokens, so \metric{} is computed against the same canonical reference as before.

\begin{figure}[t]
  \centering
  \includegraphics[width=\columnwidth]{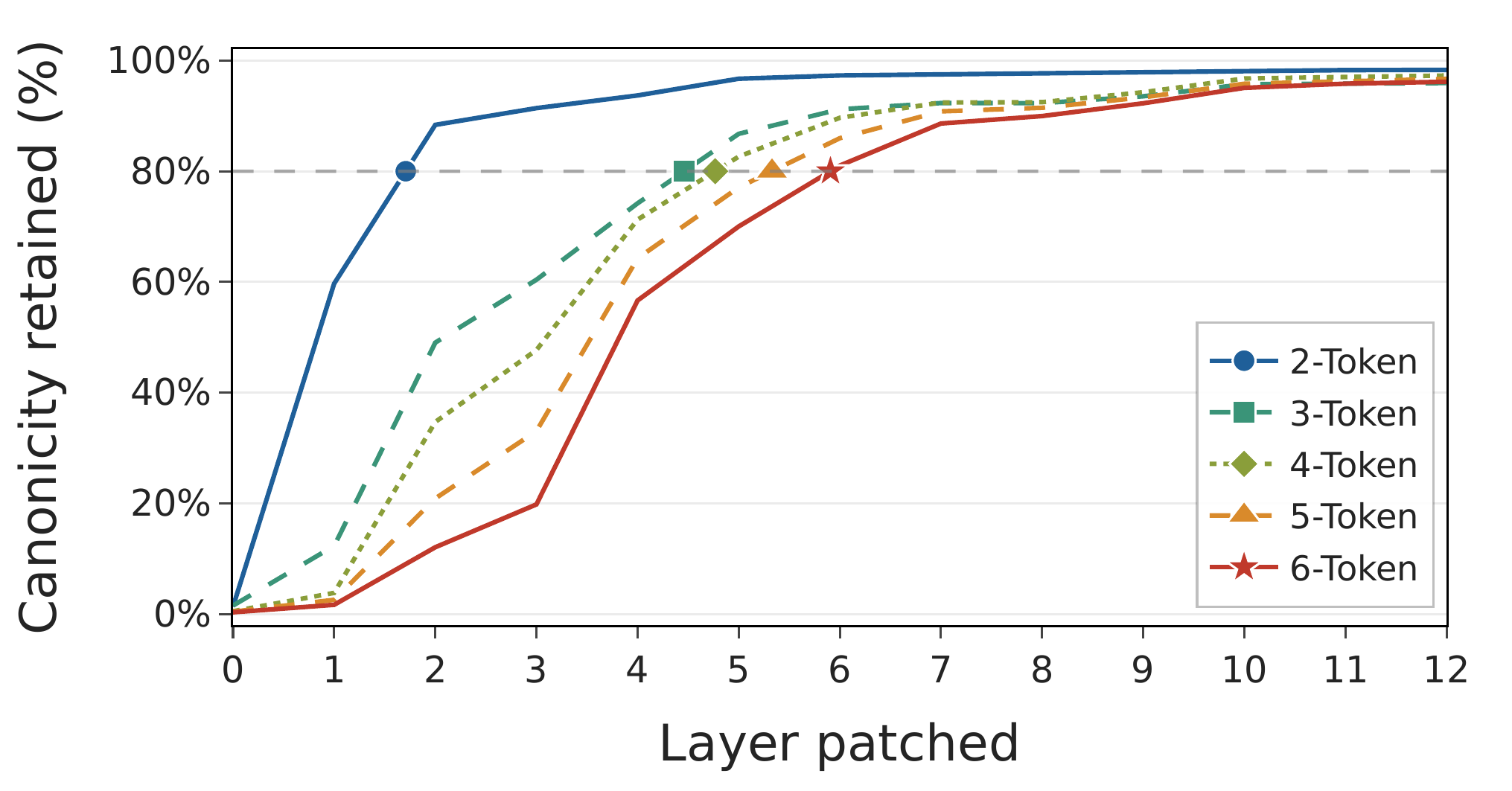}
  \caption{
  Fraction of the \metricsrc{} gap closed when patching \texttt{resid\_post}
  at the last position from a successful run into a failed one, layer by
  layer, in the first-token-differs configuration. The 2$\to$3 transition
  adds about three layers; subsequent token additions cost less than one
  layer each.}
  \label{fig:gap_closed_curves}
\end{figure}

\subsection{\Detok{} Depth Grows Sublinearly with Token Count}
\label{sec:scaling_depth}

To measure how many layers \detok{} requires at each word length, we patch \texttt{resid\_post} at the last position one layer at a time. 
A high value at layer~$l$ means the residual stream after $l$ already carries enough information to recover the canonical representation; a low value means subsequent layers are still required.

\autoref{fig:gap_closed_curves} shows the gap-closed curves.
\Detok{} is shallow at every word length: 80\% gap-closed is reached by Layer~2 for two-token words and by Layer~6 for 6-token words---19\% of the network in the deepest case.
Growth is non-uniform: the transition from two- to three-token words adds about three layers to reach the 80\% threshold, while every subsequent token requires less than one.

Word length is not the only factor.
In three-token splits where the two members share the first and last tokens and differ only in the middle one, 80\% gap-closed is reached by Layer~3, compared to Layer~5 when the first token differs (\autoref{app:token_position}).
The distance from the differing token to the last position therefore sets the depth as well.

Both findings suggest that intermediate positions contribute to the computation rather than passively passing the residual stream toward the last position.

\subsection{Intermediate Positions Relay First-Token Information}
\label{sec:relay}

To test whether intermediate positions contribute actively to \detok, we reverse the patching direction: we corrupt the successful run by replacing intermediate positions' \texttt{resid\_post} with the failed member's activations at the same position and layer.
Because the corrupted positions hold the same tokens in both members, any drop in \metric{} at the last position must come from cross-position information those positions read via attention, a direct measurement of the relay.
We report \metric{} at the last position as a percentage of the successful member's uncorrupted baseline.

\begin{figure}[t]
  \centering
  \includegraphics[width=\columnwidth]{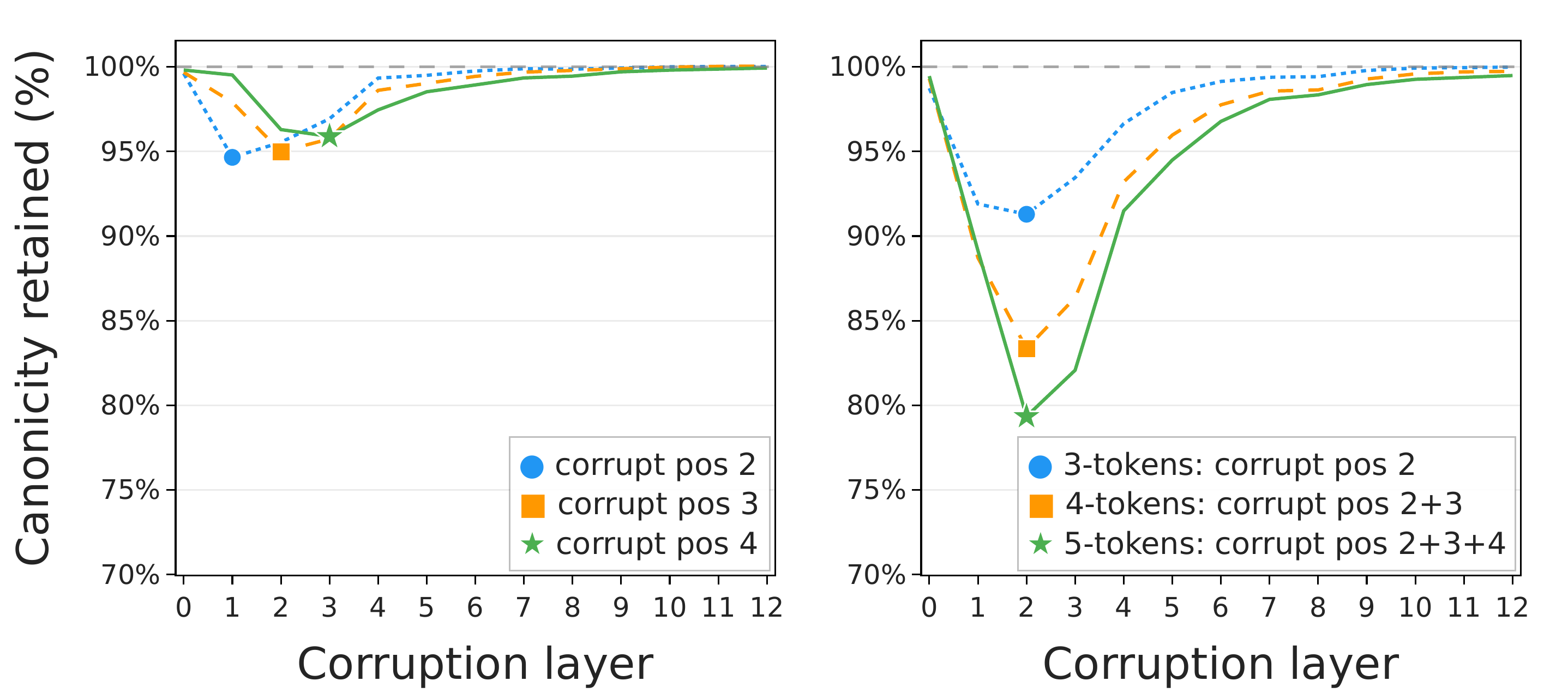}
 
  \caption{ \Metric{} at the last position after corrupting intermediate positions, relative to the same word's uncorrupted run; a larger drop indicates that the position relayed more \detok{} signal.
  \textbf{Left:} 5-token words, one position corrupted at a time; each position relays in a 2--3 layer window that shifts one layer deeper per position.
  \textbf{Right:} All intermediate positions corrupted jointly. 
  The drop deepens with word length and exceeds the sum of single-position drops (1.4--1.6$\times$), indicating partial redundancy.}

\label{fig:relay}
\end{figure}

\paragraph{Each intermediate position relays in a fixed window.}
\autoref{fig:relay} (left) shows per-position corruption curves for 5-token words. 
Each intermediate position contributes a drop concentrated in a 2--3 layer window, and these windows shift by roughly one layer per position, consistent with sequential propagation along the sequence.
The same pattern appears in 3- and 4-token words: a position's relay timing is determined by its distance from position~1, not by total word length (see \autoref{app:relay_per_position}).

\paragraph{The collective relay contribution grows with word length.}
Although each individual relay carries less signal in longer words, the collective contribution grows: corrupting all intermediate positions simultaneously produces peak drops at Layer~2: 9\% in 3-token words, rising to 21\% in 5-token words (\autoref{fig:relay}, right).
The collective drop also exceeds the sum of individual drops by an increasing factor (1.4$\times$ for 4-token, 1.6$\times$ for 5-token): each relay partially compensates when the others are intact, and this redundancy grows with the number of relay positions available.

\paragraph{}
The pattern of decreasing per-position and increasing collective contribution matches the shrinking marginal depth cost: as word length grows, intermediate positions provide parallel relay paths that limit how much each additional token costs.

%% file: 05_generalization.tex
\section{Cross-Architecture Generalization} 
\label{sec:cross_model}

\begin{figure}[t]
  \centering
  \includegraphics[width=\columnwidth]{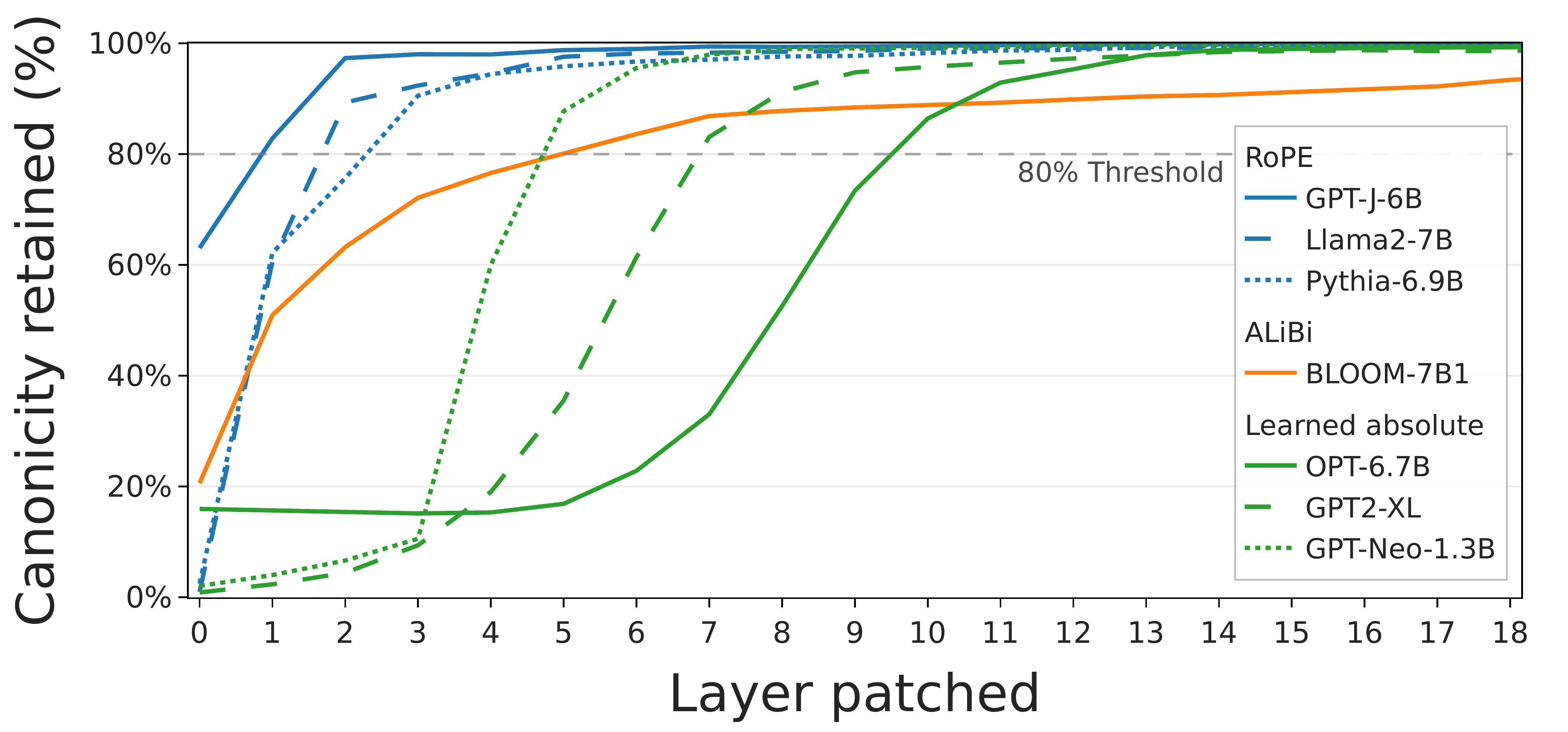}
  \caption{
  \Metric{} retained at the last subword position when patching \texttt{resid\_post} from a successful run into a failed one, in the \slt{} setting. 
  RoPE models (blue) reach high \metric{} within the first 1--5 layers; learned-absolute models (green) require 5--10 layers, accumulating the same effect gradually. 
  ALiBi (\textsc{Bloom-7B1}, orange) is intermediate.
  }
  \label{fig:cross_model_gap_closed}
\end{figure}

Does the two-stage mechanism reflect something about Llama2-7B specifically, or about how transformers compose subword tokens in general?
We extend our analysis to twelve models spanning eight families, three positional-encoding schemes (RoPE~\citep{su2021roformer}, ALiBi~\citep{press-etal-2022-train}, learned absolute~\citep{radford2019language}), and widths from 1{,}024 to 4{,}096 (\autoref{app:details}). 
We find that the two-stage process occurs in every model: attention transmits and MLP composes.
However, the depth at which it operates follows one of two distinct regimes based on a single architectural variable: \emph{positional encoding}.

\paragraph{The two-stage mechanism is universal.} 
We test each of the two stages separately across the twelve models (\autoref{app:cross_arch_two_stage_mech}). 
For attention, we patch \texttt{resid\_post} at position~1 from a successful run into a failed one, in the \slt{} setting; the layer at which this patch stops being effective marks when attention at position~2 has finished reading from position~1. 
In every model, patching position 1 closes most of the gap at the input layer, with effectiveness decaying over some early-layer window.
For the MLP, we patch \texttt{mlp\_out} from a failed run into a successful one and measure the resulting damage to canonicity; damage at a layer indicates the MLP is producing composition-specific output, not just generic local processing. 
In every model, this damage is localized to a small set of layers adjacent to the attention-read window. 
Both stages identified in \autoref{sec:mechanism} replicate in all twelve models.

\paragraph{The layer-count splits by positional encoding.}
Despite this shared structure, models differ sharply in how many layers \detok{} occupies (\autoref{fig:cross_model_gap_closed}). 
In the \emph{concentrated} regime, 80\% of the \metricsrc{} gap is closed within 1--5 layers; position~1's stream stops being essential by Layer~1--2, and MLP composition damage peaks at Layers~1--4. 
In the \emph{distributed} regime, 80\% gap-closed requires 5--10 layers; position~1 remains essential through Layers~5--8, and MLP damage peaks at mid-network. 
Positional encoding alone partitions the models: every RoPE model is concentrated and every learned-PE model is distributed, while width, depth, parameter count, tokenizer, MLP activation, and training corpus all cross the regime boundary (\autoref{app:details}).

\textsc{Bloom-7B1} (ALiBi) sits between these two poles: its processing is concentrated for 2-token words, but gradually transitions to a distributed profile as token count scales up (\autoref{app:cross_arch_gap_closed}). 
Since ALiBi and RoPE both provide relative-position access in attention, the property that aligns with the split appears to be relative-position access in general rather than rotary encoding specifically.

We hypothesize that this division exists because a relative-PE head at Layer~1 can attend from position~2 to position~1 with sharp selectivity, executing the cross-position write in one step, whereas learned-PE attention must first recover relative position from absolute coordinates, spreading the same effect over several layers; this account fits all twelve models but is untested, and we leave verification to future work.

%% file: 06_predictability.tex
\section{Early Layers Predict Success}
\label{sec:probe}
We established, via causal intervention, that \detok{} executes primarily in the early layers of every architecture tested.
This entails a falsifiable prediction: at a given depth, the residual stream should already linearly encode whether \detok{} will succeed.

We test this with a class-mean-difference probe~\citep{rimsky-etal-2024-steering, marks2024geometry}.
We rank split words by \metric{}, and take the top and bottom quintiles as \emph{successful} and \emph{failed} classes (discarding the middle 60\% to sharpen the contrast). We then fit
\[
\hat{d} = \frac{\mu_{\text{successful}} - \mu_{\text{failed}}}{\|\mu_{\text{successful}} - \mu_{\text{failed}}\|}
\]
on \texttt{resid\_post} at the final subword position in layer $l^*$, set to the gap-closed-80\% depth fixed by the mechanism (\autoref{app:cross_arch_gap_closed}).
Held-out activations are scored by projection onto $\hat{d}$; we report AUROC under 5-fold \textsc{StratifiedGroupKFold}~\citep{widodo2022stratified} grouped by canonical word.

We run this pipeline in two settings: \textbf{Isolated} uses the controlled \tokens{BOS,t1,\dots,tk} setup as before.
\textbf{In-context} embeds each split word into a WikiText-103~\citep{merity2017pointer} sentence and recomputes \metric{} per \textlangle{}word, sentence\textrangle{} record.
Within each setting, the probe is fit and evaluated on that setting's activations. 

\begin{figure}[t]
\centering
\includegraphics[width=\linewidth]{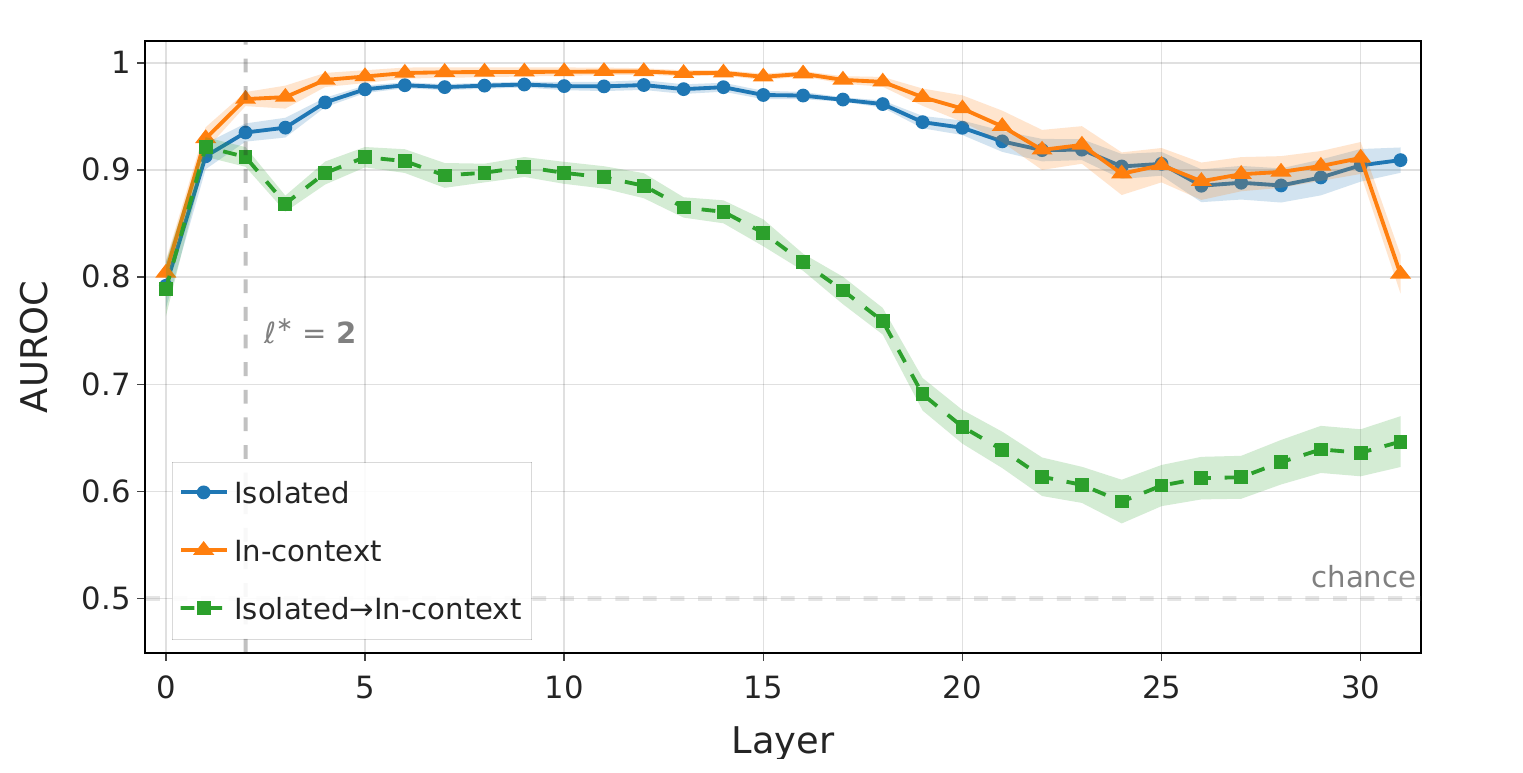}
\caption{Layerwise probe AUROC on Llama2-7B ($k{=}2$). 
Class-mean-difference probes fit on isolated activations (\textbf{Isolated}), in-context activations (\textbf{In-context}), or fit on isolated and applied to in-context activations of held-out words (\textbf{Isolated$\rightarrow$In-context}). Vertical line: $l^{*}{=}L_2$, the gap-closed-80\% depth (see \autoref{tab:models_depth_scaling_table}).}
\label{fig:llama2_probe_curves}
\end{figure}

\paragraph{The signal is linearly readable, in isolation and in context.}
On Llama2-7B at $l^{*}$=L2, Isolated achieves AUROC 0.94 and In-context 0.97 (\autoref{fig:llama2_probe_curves}).
 The probe collapses to chance under label shuffling (AUROC 0.49--0.52), confirming the signal is not an artifact of the fitting procedure. 
 The early-layer geometry thus separates both outcomes.

\paragraph{The isolated direction transfers to natural text.}
\label{cross_setting_setup}
To test if the isolated probe direction, $\hat{d}_{\text{iso}}$, generalizes to natural text, we score it on in-context activations extracted from WikiText sentences. 
To avoid lexical leakage, we apply an 80/20 train-test split to our vocabulary, stratified by isolated-setting quintiles. 
We fit $\hat{d}_{\text{iso}}$ strictly on the isolated activations of the training words, and evaluate its AUROC on the in-context activations of the held-out words.
On Llama2-7B at $l^{*}$=L2, $\hat{d}_{\text{iso}}$ achieves AUROC 0.91 on these in-context activations, within 0.03 of its own isolated test-set AUROC (0.94).
At $l^{*}$, context from preceding tokens has not yet been fully aggregated into the last subword position, so its residual still resembles the isolated counterpart.

\paragraph{The two directions align at $l^{*}$ and diverge as context accumulates.} 
At $l^*$, applying $\hat{d}_{\text{iso}}$ to in-context activations matches its own isolated test AUROC: across all 12 architectures, the gap is at most 0.04 and the median is 0.004 (\autoref{app:cross_arch_probe}). 
However, this alignment decays in deeper layers. As context aggregates into the last subword position, the in-context residual encodes outcomes in a way that $\hat{d}_{\text{iso}}$ no longer captures, causing its AUROC to drop well below the In-context probe. 
This is consistent with the mechanism: 
$\hat{d}_{\text{iso}}$ is fit where the last position carries composition alone, which is no longer the sole signal once context accumulates.

\paragraph{Cross-architecture replication.} 
The same construction yields Isolated AUROC 0.76--0.94 and In-context 0.86--0.99 at $l^*$ across all 12 models (\autoref{app:cross_arch_probe}). 
The two regimes of \S\ref{sec:cross_model} appear in the layer-by-layer profile: concentrated-regime models reach peak AUROC within the first 10--15\% of layers; distributed-regime models build up gradually over the first third. 
Causal intervention and correlational probing thus identify the same critical depth $l^*$, supporting a single underlying mechanism.

%% file: 07_conclusion.tex
\section{Conclusion}
\label{sec:conclusion}

We characterized \detok{} in transformer language models as an early-layer two-stage mechanism: attention writes a token-specific directional signal from preceding subwords, and the MLP composes it with the local embedding.
The mechanism is consistent across twelve models from eight families and extends to longer words via sequential relays at intermediate positions, with its depth governed by positional encoding.

Beyond mapping the phenomenon, we established that the \detok{} outcome is linearly readable from early layers. 
This opens the door to practical, inference-time interventions, such as flagging tokenization-induced failures before text generation concludes. 
Furthermore, our findings suggest that early-layer activations can be leveraged as an intrinsic metric for tokenizer evaluation, offering a clean, mechanistic alternative to behavioral criteria that inherently conflate tokenizer and model performance.

%% file: 08_limitations.tex
\section*{Limitations}

\paragraph{\Metric{} measures alignment to a model-internal reference.}
Our success metric quantifies how closely a split word's hidden state aligns with the model's own hidden state for the canonical single-token form of the same word (\S\ref{sec:metric}).
This reference is itself a representation the model computes, not an external ground truth for the word's meaning.
A split input whose hidden state diverges from the canonical could in principle still support coherent downstream computation through a different geometry, and would be scored as a partial failure under our metric.
Consistent with this, the behavioral agreement in \S\ref{sec:metric} is strong and monotonic, but not absolute.
The mechanism we identify is therefore the one that produces canonical-aligned representations. 
Whether transformers also implement alternative pathways that produce a word-level representation through different geometry remains outside the scope of our evidence.

\paragraph{Restriction to single-token words.}
Our paired design requires a canonical single token form to serve as the reference, which forces us to study only words the tokenizer can represent as a single token, artificially split into multi-token sequences.
The words for which \detok{} arguably matters most behaviorally (rare words, technical terms, and proper names that the tokenizer represents as two or more tokens) have no single-token counterpart and therefore no comparable ground truth available to our method.
Whether the two-stage mechanism we identify governs these naturally-multi-token words is an open question we cannot address with the present methodology, and would require either a different reference signal or an alternative behavioral criterion for \detok{} success.

%% file: 09_appendix.tex
\input{appendix/data_source}

\input{appendix/cross_arch_metric_layer}

\input{appendix/detok_specificity}

\input{appendix/fst_attn_similarity.tex}


\input{appendix/relay_per_position}

\input{appendix/models_details}

\input{appendix/cross_architecture_mechanism}

\input{appendix/gap_closed_per_model}

\input{appendix/cross_arch_probe}


%% file: appendix/data_source.tex
\section{Data Sources and Single-Token Word Pools}
\label{app:data_sources}
All experiments operate on a per-model pool of \emph{single-token words}: English words that the model's tokenizer encodes as exactly one token (excluding any special tokens). 
Restricting to such words is necessary because our analysis compares a split tokenization of a word against its canonical single-token representation (\autoref{sec:metric}); inherently multi-token words admit no such reference.

\paragraph{Raw word lists.}
We draw candidate words from three publicly available English word lists:
\begin{itemize}
    \item A frequency-ordered list of the 100{,}000 most common lowercased English words from the \texttt{top-english-wordlists} repository.\footnote{\url{https://github.com/david47k/top-english-wordlists}}
    \item The Google 10{,}000-word common English list.\footnote{\url{https://github.com/first20hours/google-10000-english}}
    \item The English word frequency dataset of \citet{tatman2017englishfreq}, a unigram frequency table of approximately 333{,}000 (word, count) entries derived from the Google Web Trillion Word Corpus.\footnote{\url{https://www.kaggle.com/datasets/rtatman/english-word-frequency}}
\end{itemize}

For both the contrastive-pair datasets used in activation patching
(\autoref{sec:mechanism} onward) and the behavioral readout-layer
validation (\autoref{sec:metric}), we draw on all three lists.
We concatenate the three lists and merge them into a single set of
candidate words.
The merged words are lowercased and deduplicated, yielding a combined
candidate pool of 338{,}850 unique words.
Each candidate is then tokenized with the model's own tokenizer, and
only words encoded as exactly one content token are retained, giving
the per-model pool.

\paragraph{Per-model single-token filtering.}
Because tokenizer vocabularies differ substantially across architectures, the resulting pool size varies: from roughly 7.5k words for GPT-2--family models to roughly 28k for Gemma-2-2B (SentencePiece Unigram).
Pool sizes per model and the resulting number of contrastive pairs are listed in \autoref{tab:pool_sizes}.

\begin{table}[t]
\centering
\small
\begin{tabular}{l r r r}
\toprule
Model & Pool & Threshold & 2-token pairs \\
\midrule
GPT-2 Large   & 7{,}521  & 0.15 & 4{,}418 \\
GPT-2 XL      & 7{,}521  & 0.20 & 4{,}918 \\
GPT-Neo 1.3B  & 7{,}521  & 0.05 & 4{,}407 \\
GPT-J 6B      & 7{,}521  & 0.15 & 4{,}791 \\
OPT 1.3B      & 7{,}521  & 0.20 & 12{,}376 \\
OPT 6.7B      & 7{,}521  & 0.20 & 11{,}207 \\
Pythia 410M   & 8{,}651  & 0.15 & 8{,}983 \\
Pythia 1B     & 8{,}651  & 0.15 & 7{,}997 \\
Pythia 6.9B   & 8{,}651  & 0.10 & 7{,}499 \\
Llama-2-7B    & 8{,}933  & 0.30 & 12{,}663 \\
Gemma-2-2B    & 28{,}006 & 0.30 & 9{,}954 \\
BLOOM-7B1    & 8{,}933  & 0.30 & 9{,}961 \\
\bottomrule
\end{tabular}
\caption{Per-model dataset sizes. 
\textbf{Pool}: unique words that the model's tokenizer encodes as exactly one token, after merging the three source lists, lowercasing, and deduplication. 
\textbf{Threshold}: per-model minimum metric gap $|\Delta\metric|$ required to retain a pair, calibrated to cross-architecture variation in cosine-similarity spread~\citep{ethayarajh-2019-contextual}. 
\textbf{2-token pairs}: contrastive pairs used in activation patching, summed across the \slt{} and \sft{} settings.}

\label{tab:pool_sizes}
\end{table}

%% file: appendix/cross_arch_metric_layer.tex
\section{Choosing the Readout Layer and Validating \Metric{}}
\label{app:metric_validation}

This appendix supports two claims from \S\ref{sec:metric}: that layer $n-2$ is the right readout depth, and that \metric{} at that depth tracks model behavior.

\paragraph{Setup.}
For each model we use the flat pool of all valid two-token
segmentations of its single-token words, built from the per-model pool
described in Appendix~\ref{app:data_sources}.
For each segmentation we run two forward passes, one on the split input and one on the canonical input.
At every layer $L$ we record the cosine similarity between their last-position residual streams.
At the last token position of each input we also record the next-token distribution and compute three behavioral comparisons between the split and canonical predictions: $\mathrm{KL}(p_{\text{canon}} \,\|\, p_{\text{split}})$, top-1 agreement (whether the two inputs share the same argmax next token), and top-5 overlap.
At each layer we then compute the Spearman correlation across all rows between the cosine similarity and each behavioral metric.

\begin{figure*}
    \includegraphics[width=1\linewidth]{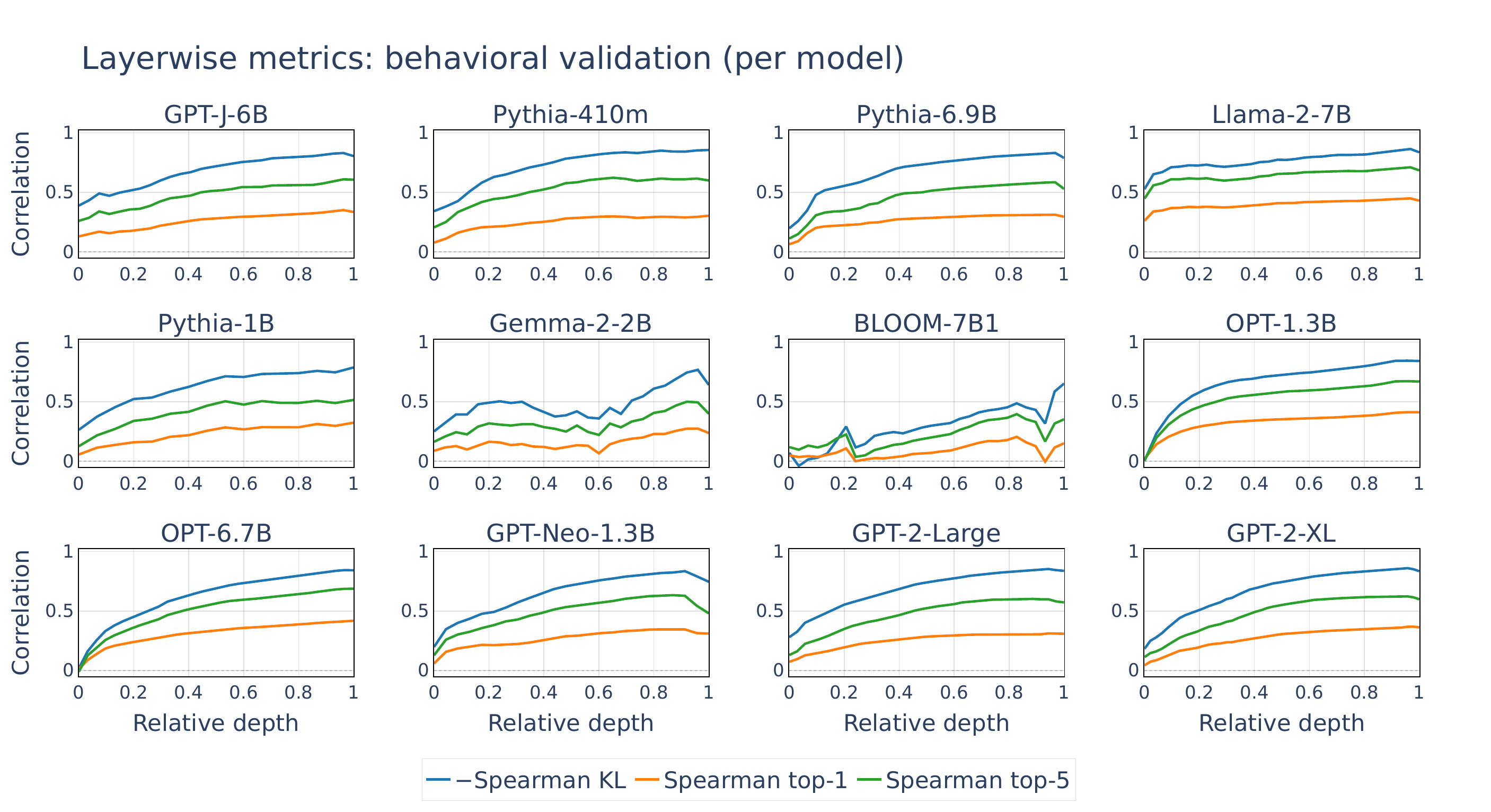}
    \caption{Per-layer Spearman correlation between cosine similarity
    to the canonical representation and three next-token behavioral
    metrics---KL divergence (blue), top-1 agreement (orange), and top-5
    overlap (green)---across 12 architectures. 
    The x-axis is relative depth; KL correlation is plotted as $-\rho_{\text{KL}}$ so that higher is better on all three curves. 
    Across every model, all three correlations rise with depth and peak near relative depth $\sim 0.95$ (the $n-2$ layer used as the measurement point in \S\ref{sec:metric}), then drop at the final layer due to next-token-prediction reshaping. BLOOM-7B1 shows noisier curves but the same overall shape.}
    \label{fig:cross_arch_metric_layer}
\end{figure*}

\paragraph{Layer selection.}
\autoref{fig:cross_arch_metric_layer} shows the three per-layer correlations across all 12 architectures.
In every model, all three correlations rise with depth and peak near layer $n{-}2$, then drop sharply at the final layer due to next-token-prediction reshaping~\citep{belrose2023eliciting}.
On Llama2-7B, layer 30 yields $\rho_{\text{KL}} = -0.86$, $\rho_{\text{top-1}} = +0.45$, and $\rho_{\text{top-5}} = +0.71$, ranking first among all 32 layers.
For the remaining models, $n-2$ ranks first or within a negligible margin of the top-ranked layer, consistently under all three behavioral metrics.
We therefore adopt $L = n{-}2$ as the readout layer throughout the paper.

\begin{figure}[t]
  \centering
  \includegraphics[width=\columnwidth]{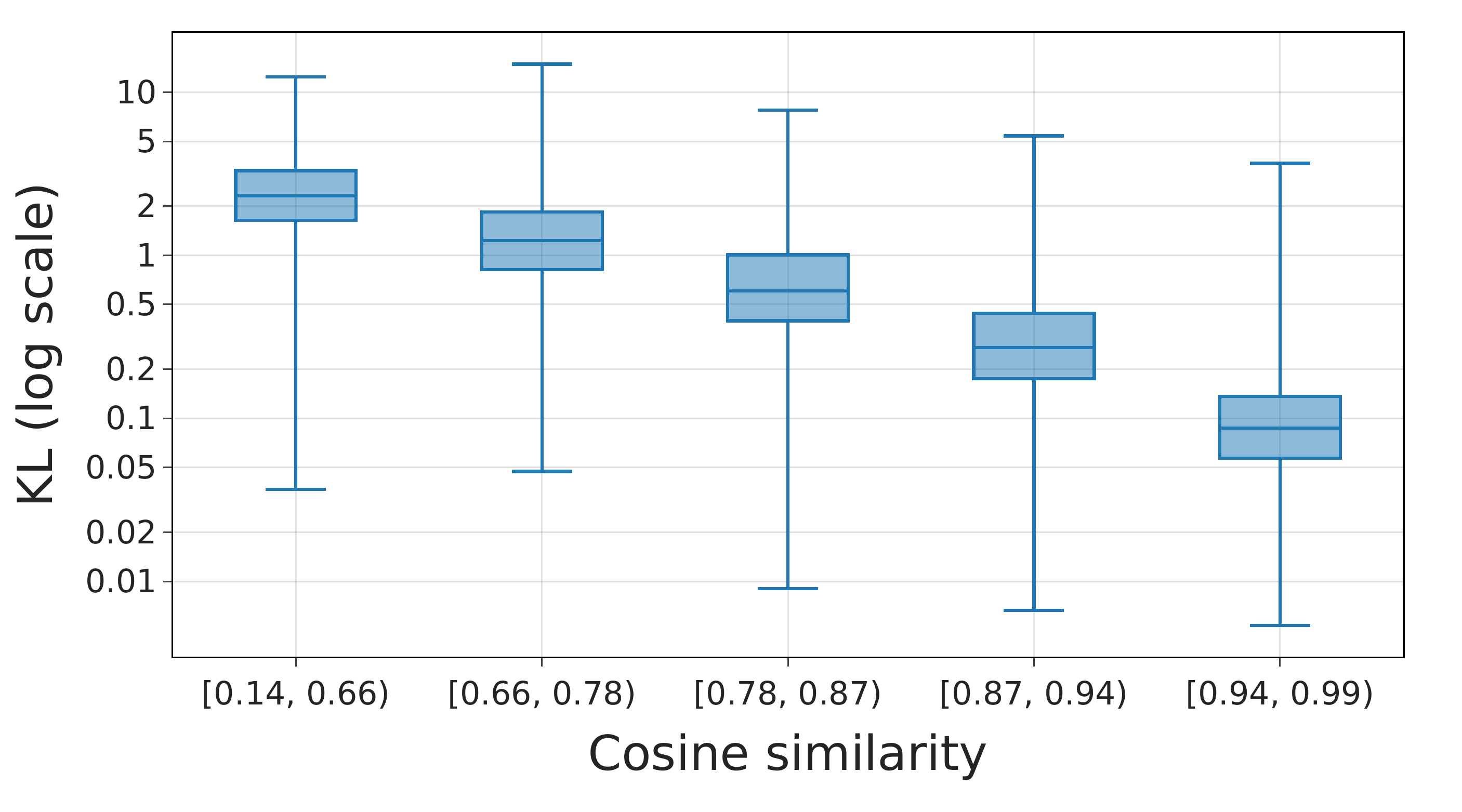}
  \caption{Next-token KL divergence between split-input and canonical-input distributions, binned by \metric{} (y-axis log scale). Median KL falls monotonically across bins, dropping roughly $25\times$ from the lowest to the highest \metric{} bin.}
  \label{fig:metric_validation}
\end{figure}

\paragraph{Behavioral validation.}
Having fixed the readout layer, we verify that \metric{} at that layer predicts behavior rather than reflecting a geometric artifact.
We rank words by \metric{} and stratify them into quintiles.
\Metric{} is a strong, monotonic predictor of behavioral agreement: words in the top quintile (mean \metric{} 0.97) reach 78\% top-1 agreement with the canonical input and a KL divergence of 0.09, while words in the bottom quintile (mean \metric{} 0.65) reach only 24\% top-1 agreement and a KL divergence of 1.86.
\autoref{fig:metric_validation} shows the full relationship: median KL falls roughly $25\times$ from the bottom to the top of the \metric{} range, and the trend is monotonic across all bins.

\paragraph{Qualitative examples.}
\S\ref{sec:metric} gives one example of what a low \metric{} value costs. \autoref{tab:qualitative_examples} gives five more.
Each block holds one word, its canonical single-token input, and two splits of that word, one with high \metric{} and one with low.
Word identity is held constant, so any difference in behavior comes from the segmentation alone.
The high-\metric{} split reproduces the canonical continuations in every case, while the low-\metric{} split shares no token with the canonical top-5 prediction in four of the five.
The failed predictions are not noise.
In several cases the continuation completes the final subword into a different word: \tokens{n} becomes \say{nights} for \tokens{\_show,n} and \say{nation} for \tokens{\_media,n}, just as \tokens{d} becomes \say{draft} and \say{diet} for \tokens{\_advance,d} in \S\ref{sec:metric}.
A low \metric{} value therefore means that the model reads the last subword as the start of a new word, rather than as the end of the intended one.

\begin{table*}[t]
\centering
\small
\setlength{\tabcolsep}{5pt}
\begin{tabular}{llccl}
\toprule
Input & \metric{} & KL$\downarrow$ & top-5 & Top-3 next tokens (\%) \\
\midrule
\tokens{\_corpor}   & ---  & ---  & ---  & \texttt{ate}~82.9 \textbar{} \texttt{ation}~6.0 \textbar{} \texttt{ations}~4.1 \\
\tokens{\_cor,por}  & 0.95 & 0.01 & 1.00 & \texttt{ate}~84.9 \textbar{} \texttt{ation}~6.1 \textbar{} \texttt{ations}~2.6 \\
\tokens{\_corpo,r}  & 0.33 & 9.54 & 0.00 & \texttt{ison}~9.8 \textbar{} \texttt{ub}~7.6 \textbar{} \texttt{re}~4.6 \\
\midrule
\tokens{\_shown}    & ---  & ---  & ---  & \texttt{in}~21.1 \textbar{} \texttt{on}~8.8 \textbar{} \texttt{below}~8.2 \\
\tokens{\_sh,own}   & 0.96 & 0.06 & 0.80 & \texttt{in}~19.6 \textbar{} \texttt{on}~7.2 \textbar{} \texttt{below}~7.2 \\
\tokens{\_show,n}   & 0.29 & 4.51 & 0.00 & \texttt{ights}~4.1 \textbar{} \texttt{roll}~3.2 \textbar{} \texttt{eth}~2.7 \\
\midrule
\tokens{\_median}   & ---  & ---  & ---  & \texttt{household}~10.6 \textbar{} \texttt{income}~6.1 \textbar{} \texttt{sal}~5.7 \\
\tokens{\_med,ian}  & 0.92 & 0.31 & 0.40 & \texttt{et}~8.5 \textbar{} \texttt{income}~5.2 \textbar{} \texttt{of}~3.8 \\
\tokens{\_media,n}  & 0.26 & 5.85 & 0.00 & \texttt{ation}~15.9 \textbar{} \texttt{in}~4.6 \textbar{} \texttt{ama}~3.3 \\
\midrule
\tokens{\_gender}   & ---  & ---  & ---  & \texttt{,}~10.8 \textbar{} \texttt{and}~6.5 \textbar{} \texttt{equality}~6.1 \\
\tokens{\_g,ender}  & 0.98 & 0.05 & 1.00 & \texttt{,}~7.0 \textbar{} \texttt{-}~6.6 \textbar{} \texttt{equality}~5.1 \\
\tokens{\_ge,nder}  & 0.36 & 3.85 & 0.20 & \texttt{8}~8.3 \textbar{} \texttt{k}~6.1 \textbar{} \texttt{-}~4.5 \\
\midrule
\tokens{\_cried}    & ---  & ---  & ---  & \texttt{out}~13.4 \textbar{} \texttt{,}~10.4 \textbar{} \texttt{the}~10.4 \\
\tokens{\_c,ried}   & 0.92 & 0.27 & 0.80 & \texttt{out}~14.8 \textbar{} \texttt{,}~6.6 \textbar{} \texttt{the}~4.0 \\
\tokens{\_cr,ied}   & 0.29 & 4.26 & 0.00 & \texttt{ie}~41.8 \textbar{} \texttt{it}~3.9 \textbar{} \texttt{e}~3.0 \\
\bottomrule
\end{tabular}
\caption{Next-token behavior for two splits of the same word, scored against the canonical single-token input.
Each block gives the canonical input, the highest-\metric{} split, and the lowest-\metric{} split.
KL is $\mathrm{KL}(p_{\text{canon}}\|\cdot)$, and top-5 is the overlap with the canonical top-5 prediction.
The five words are drawn from those with the largest gap between their best and worst split.}
\label{tab:qualitative_examples}
\end{table*}

\subsection{Causal Behavioral Validation}
\label{app:causal_behavior_validation}
The validation above is observational: \metric{} correlates with next-token behavior across words. 
Here we confirm the same relationship causally. 
We take the \slt{} L1 \texttt{attn\_out} patch and the \sft{} L1 \texttt{mlp\_out} patch, the interventions that localize the two stages of the mechanism in \S\ref{sec:mechanism}, and ask whether they move model behavior and not only \metric{}.

Rather than scoring \metric{}, we score the patched run's next-token distribution against the canonical single-token run, using top-1 agreement, top-5 overlap, and $\mathrm{KL}(p_{\text{canon}}\,\|\,\cdot)$. 
As in \S\ref{sec:patching_method}, we report the fraction of the gap closed from the failed split to the successful split: 0\% is the unpatched \metricsrc{} score of failed split, and 100\% is the unpatched \metric{} score of the successful split. 
As a control (\textsc{Random}), we run the same intervention but take the patched activation from an unrelated word at the same component, layer, and position, rather than from the pair's successful member. 
For example, in \slt{} the failed split \tokens{\_err, or} is normally patched at L1 \texttt{attn\_out}, position~2, with the activation of its successful partner \tokens{\_po, or}; the \textsc{Random} control instead patches the same site with the L1
\texttt{attn\_out} activation, at its own last position, of an unrelated word such as \tokens{\_ta, ble}.

\autoref{tab:causal_behavior_table} shows that both interventions shift every behavioral metric from the failed split toward the successful one. 
The L1 \texttt{attn\_out} patch raises top-1 agreement with the canonical prediction from 3.5\% to 27.6\%, closing 42\% of the gap against only 1\% for the \textsc{Random} control, and the L1 \texttt{mlp\_out} patch reproduces this (1.9\% to 26.8\%, 40\% gap closed against 2\%). 
Across metrics the \texttt{attn\_out} patch closes 42\% of the top-1 gap and 46\% of top-5, with KL closing 60\%; the \metricsrc{} column produces the 53\%--the gap stated in \S\ref{sec:mechanism}. 
The behavioral effect is thus of the same order as the representational one, substantial on every metric, confirming that the intervention moves model behavior, not only \metric{}. 
This is a one-time check on the metric; all other results in the paper are reported in \metric{} alone.

\begin{table}[t]
\centering
\footnotesize
\setlength{\tabcolsep}{1.6pt}
\begin{tabular}{lcccc}
\toprule
& top-1 & top-5 & KL$\downarrow$ & \texttt{\metricsrc{}} \\
\midrule
\multicolumn{5}{l}{\textit{\slt{}:} L1 \texttt{attn\_out} patch} \\
\midrule
Failed split       & 3.5\%  & 13.1\% & 3.75 & 0.44 \\
\textsc{Random}    & 3.9\%  & 13.8\% & 3.49 & 0.48 \\
\textsc{Patched}   & 27.6\% & 39.6\% & 1.76 & 0.69 \\
Successful split   & 60.4\% & 70.4\% & 0.43 & 0.91 \\
\addlinespace
Gap closed (\textsc{Patched}) & 42\% & 46\% & 60\% & 53\% \\
Gap closed (\textsc{Random})  & \phantom{0}1\% & \phantom{0}1\% & \phantom{0}8\% & \phantom{0}8\% \\
\midrule
\multicolumn{5}{l}{\textit{\sft{}:} L1 \texttt{mlp\_out} patch} \\
\midrule
Failed split       & 1.9\%  & 10.7\% & 4.14 & 0.41 \\
\textsc{Random}    & 2.9\%  & 14.4\% & 3.68 & 0.47 \\
\textsc{Patched}   & 26.8\% & 38.6\% & 1.89 & 0.68 \\
Successful split   & 64.0\% & 74.1\% & 0.30 & 0.93 \\
\addlinespace
Gap closed (\textsc{Patched}) & 40\% & 44\% & 59\% & 52\% \\
Gap closed (\textsc{Random})  & \phantom{0}2\% & \phantom{0}6\% & 12\% & 12\% \\
\bottomrule
\end{tabular}
\caption{Behavioral effect of the two interventions of
\S\ref{sec:mechanism}, scored against the canonical single-token run. top-1
and top-5 are agreement rates with the canonical next-token prediction; KL
is $\mathrm{KL}(p_{\text{canon}}\|\cdot)$, lower is better;
\metricsrc{} at $n{-}2$, shown for comparison.\ \textsc{Random} patches from an
unrelated word.}
\label{tab:causal_behavior_table}
\end{table}

%% file: appendix/detok_specificity.tex
\section{Specificity to \Detok{}}
\label{app:detok_specificity}

Layer~1 attention reads from the preceding position in ordinary text as well as in split words.
This appendix asks whether the effect of \S\ref{sec:patching} is specific to \detok{}, or a generic property of early-layer attention.

We construct 1{,}000 sequences of the form \tokens{BOS, w1, w2}, where both words are single-token English words, and pair them so that the two members share \tokens{w2} and differ in \tokens{w1} (e.g., \tokens{BOS, the, cat} vs.\ \tokens{BOS, a, cat}).
This is the structural analogue of our \slt{} setting (\texttt{\_po|or} vs.\ \texttt{\_err|or}), with two words in place of two subwords of one word.
We then apply the identical intervention: patching Layer~1 \texttt{attn\_out} at position~2 from the source run into the target run.

A two-word sequence has no canonical single-token reference, so \metric{} is undefined for it.
We therefore score both conditions by alignment to the source run's own layer-30 representation at the last position.
Under this metric the intervention closes 50.8\% of the gap for artificially split words, and 6.2\% for ordinary two-word sequences.

Layer~1 attention therefore takes part in general cross-token processing, but its causal contribution to the downstream representation is roughly eight times larger during \detok{}.

%% file: appendix/fst_attn_similarity.tex
\section{Attention Output Similarity in \sft{} Pairs}
\label{app:fst_attn_similarity}

\S\ref{sec:\sft_asymmetry} states that Layer~1 \texttt{attn\_out} at position~2 is nearly identical within an \sft{} pair.
This does not follow from the pair construction.
Sharing the first token fixes the value vector at position~1, but the second-token embedding differs, so the query at position~2, and therefore the attention pattern, may differ as well.
We measure the resulting similarity directly.

For each of the 6{,}254 \sft{} pairs we compute the cosine similarity between the source and target L1 \texttt{attn\_out} at position~2.
As a baseline we repeat the computation against words with a different first token.
\autoref{tab:fst_attn_similarity} reports both, for the full attention output and for the three heads that carry the \detok{} signal (\S\ref{sec:ov_circuit}).

The per-head numbers show what drives the similarity.
The baseline falls from 0.739 for the full output to 0.144--0.231 per head.
Within a pair, by contrast, the three heads stay at 0.834--0.998.
The high within-pair similarity is therefore a consequence of the shared first token, not of a direction these heads write for every input.
Head~27, which alone closes 17\% of the gap in \S\ref{sec:ov_circuit}, is the most nearly identical of the three at 0.998.

\begin{table}[t]
\centering
\small
\begin{tabular}{lcc}
\toprule
Component & Within pair & Different first token \\
\midrule
\texttt{attn\_out} & 0.948 & 0.739 \\
\midrule
Head 24 & 0.923 & 0.214 \\
Head 27 & 0.998 & 0.144 \\
Head 28 & 0.834 & 0.231 \\
\bottomrule
\end{tabular}
\caption{Cosine similarity of Layer~1 \texttt{attn\_out} at position~2 across all 6{,}254 \sft{} pairs on Llama2-7B.
\textbf{Within pair}: source against target. \textbf{Different first token}: against words that do not share the first token.}
\label{tab:fst_attn_similarity}
\end{table}

%% file: appendix/relay_per_position.tex
\section{The Effect of Token Position on \Detok{} Depth}
\label{app:token_position}

\S\ref{sec:scaling} measures \detok{} depth as word length grows.
This appendix holds length fixed and varies position instead, in two ways: by moving the differing token within the word, and by timing the contribution of each intermediate position.

\subsection{Depth for Three-Token Splits with a Differing Middle Token}
\label{app:middle_token_patching}

We construct 4{,}609 three-token pairs on Llama2-7B in which the two members share the first and the last token and differ only in the middle one.
Apart from the position of the differing token, the construction follows \S\ref{sec:datasets}.
We patch \texttt{resid\_post} at the last position, layer by layer, as in \autoref{fig:gap_closed_curves}.

The curve reaches 80\% gap-closed at Layer~3.
Three-token splits whose first token differs reach the same threshold at Layer~5, and two-token splits reach it at Layer~2.
Depth therefore tracks the distance from the differing token to the last position, and not the token count alone: a differing middle token starts one position closer to the readout, and costs about two layers less.
\autoref{app:relay_per_position} shows the same dependence from the other direction.

\subsection{Relay Timing per Intermediate Position in 3- and 4-Token Splits}
\label{app:relay_per_position}

\S\ref{sec:relay} establishes, on 5-token words, that each intermediate position relays first-token information in a fixed 2--3 layer window whose timing is set by the position's index.
This subsection verifies that the same per-position timing holds for shorter words.

\begin{figure}
    \includegraphics[width=\columnwidth]{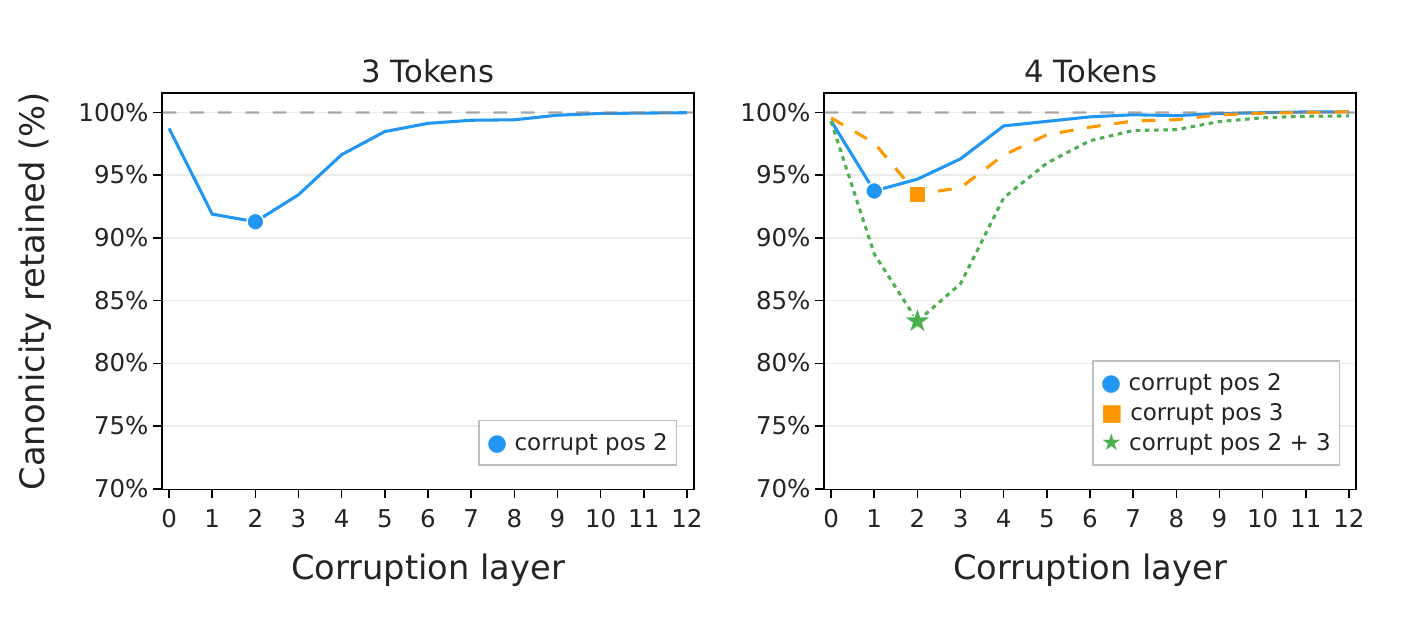}
    \caption{Per-position relay across word lengths.
    \Metric{} at the last position when a single intermediate position's \texttt{resid\_post} is patched with the failed run's activations at one layer at a time, as a percentage of the uncorrupted baseline.
    \textbf{Left:} 3-token words (one intermediate position: 2).
    \textbf{Right:} 4-token words (intermediate positions: 2, 3).
    Across both lengths relay timing is set by a position's index, not by total word length.}
    \label{fig:relay_per_ntokens_per_position}
\end{figure}

We repeat the single-position corruption analysis of \autoref{fig:relay} on 3- and 4-token words, corrupting each intermediate position's \texttt{resid\_post} at one layer at a time and measuring the drop in \metric{} at the last position.
\autoref{fig:relay_per_ntokens_per_position} shows the result.
The 3-token case has a single intermediate position (position~2), which drops \metric{} maximally at Layer~2---the same layer at which position~2 drops \metric{} in 5-token words (\autoref{fig:relay}, left). 
The 4-token case adds position~3, which contributes its own drop one layer later, again matching its 5-token counterpart. 
A position's relay timing is therefore a function of its index alone; total word length affects the \emph{number} of relays available but not \emph{when} each one activates.

%% file: appendix/models_details.tex
\section{Model Details}
\label{app:details}

\begin{table*}[t]
\centering
\small
\setlength{\tabcolsep}{2pt}
\begin{tabular}{lcccccclccc}
\toprule
\textbf{Model} & \textbf{Family} & \textbf{Layers} & $\boldsymbol{d_\text{model}}$ & \textbf{Params} & \textbf{Training data} & \textbf{Tokenizer} & \textbf{MLP} & \textbf{PE} & \textbf{L1 (\%)} & \textbf{L@80\%} \\
\midrule
\multicolumn{11}{l}{\textit{Concentrated}} \\
GPT-J-6B      & GPT-J   & 28 & 4{,}096 & 6.0B  & The Pile    & GPT-2 BPE   & GELU   & RoPE  & 82.8 & 1 \\
Pythia-410M   & Pythia  & 24 & 1{,}024 & 410M  & The Pile    & GPT-NeoX BPE & GELU   & RoPE  & 74.0 & 3 \\
Pythia-6.9B   & Pythia  & 32 & 4{,}096 & 6.9B  & The Pile    & GPT-NeoX BPE & GELU   & RoPE  & 62.2 & 3 \\
Llama2-7B     & LLaMA   & 32 & 4{,}096 & 7.0B  & Mixed web   & SP BPE & SwiGLU & RoPE  & 60.5 & 2 \\
Pythia-1B     & Pythia  & 16 & 2{,}048 & 1.0B  & The Pile    & GPT-NeoX BPE & GELU   & RoPE  & 59.9 & 3 \\
Gemma-2-2B    & Gemma 2 & 26 & 2{,}304 & 2.6B  & Mixed web   & SP Unigram & GeGLU  & RoPE  & 51.9 & 2 \\
\midrule
\multicolumn{11}{l}{\textit{Intermediate (ALiBi)}} \\
Bloom-7B1     & BLOOM   & 30 & 4{,}096 & 7.1B  & ROOTS       & BLOOM BPE & GELU   & ALiBi & 51.0 & 5 \\
\midrule
\multicolumn{11}{l}{\textit{Distributed}} \\
OPT-1.3B      & OPT     & 24 & 2{,}048 & 1.3B  & Mixed web   & GPT-2 BPE & ReLU   & Learned & 25.8 & 7 \\
OPT-6.7B      & OPT     & 32 & 4{,}096 & 6.7B  & Mixed web   & GPT-2 BPE & ReLU   & Learned & 15.7 & 10 \\
GPT-Neo-1.3B  & GPT-Neo & 24 & 2{,}048 & 1.3B  & The Pile    & GPT-2 BPE & GELU   & Learned & 4.0  & 5 \\
GPT2-Large    & GPT-2   & 36 & 1{,}280 & 774M  & WebText     & GPT-2 BPE & GELU   & Learned & 2.3  & 7 \\
GPT2-XL       & GPT-2   & 48 & 1{,}600 & 1.5B  & WebText     & GPT-2 BPE & GELU   & Learned & 2.3  & 7 \\
\bottomrule
\end{tabular}
\caption{Architectural details and patching results for all 12 models, 2-token words, \slt{} configuration.
\textbf{L1 (\%)}: percentage of \metricsrc{} gap closed by
patching Layer-1 \texttt{resid\_post}. 
\textbf{L@80\%}: first layer at which $\geq 80\%$ of the gap is closed. 
Models are sorted within each group by L1. SP = SentencePiece.}
\label{tab:models_details_table}
\end{table*}

\autoref{tab:models_details_table} lists the 12 models analyzed in this work, grouped by the regime identified in \S\ref{sec:cross_model}. 
We draw from eight families spanning three positional-encoding schemes: GPT-2~\citep{radford2019language}, GPT-Neo~\citep{black2021gptneo}, GPT-J~\citep{wang2021gptj}, OPT~\citep{zhang2022opt}, Pythia~\citep{biderman2023pythia}, BLOOM~\citep{workshop2022bloom}, Llama~2~\citep{touvron2023llama2}, and Gemma~2~\citep{team2024gemma}. 
The table supports the claim that among the architectural variables we vary, positional encoding is the only one that aligns with the regime split.
Every other column crosses the regime boundary. 
\emph{Depth} ranges from 16 to 32 layers within the concentrated regime alone (Pythia-1B to Pythia-6.9B and Llama2-7B), and from 24 to 48 within distributed (OPT-1.3B to GPT2-XL); the ranges overlap
substantially. 
\emph{Width} ($d_\text{model}$) similarly spans 1{,}024--4{,}096 in concentrated and 1{,}280--4{,}096 in distributed. 
\emph{Parameter count} spans 410M--7.1B in concentrated and 774M--6.7B in distributed---nearly identical ranges. 
\emph{Training corpus} crosses the boundary in both directions: The Pile trains concentrated-regime models (Pythia family, GPT-J-6B) and a distributed-regime model (GPT-Neo-1.3B); mixed web data trains concentrated-regime models (Llama2-7B, Gemma-2-2B) and distributed-regime models
(OPT-1.3B, OPT-6.7B). 
\emph{Tokenizer} crosses the boundary as well: GPT-2 BPE appears in GPT-J-6B on the concentrated side and in all five distributed models. 
\emph{MLP activation} likewise crosses: GELU is used by most concentrated models and by three
of the five distributed ones.

\textsc{Bloom-7B1} (ALiBi) sits between the two regimes, concentrated at 2-token words but drifting toward distributed depth as token count grows (\autoref{app:cross_arch_gap_closed}). 
This intermediate behavior is itself informative: ALiBi and RoPE are mechanistically distinct schemes that both provide relative-position access in attention, so the property aligned with the split appears to be relative-position access in general rather than rotary encoding specifically.

%% file: appendix/cross_architecture_mechanism.tex
\section{Cross-Architecture Two-Stage Mechanism}
\label{app:cross_arch_two_stage_mech}

\begin{figure*}
    \centering
    \includegraphics[width=1\linewidth]{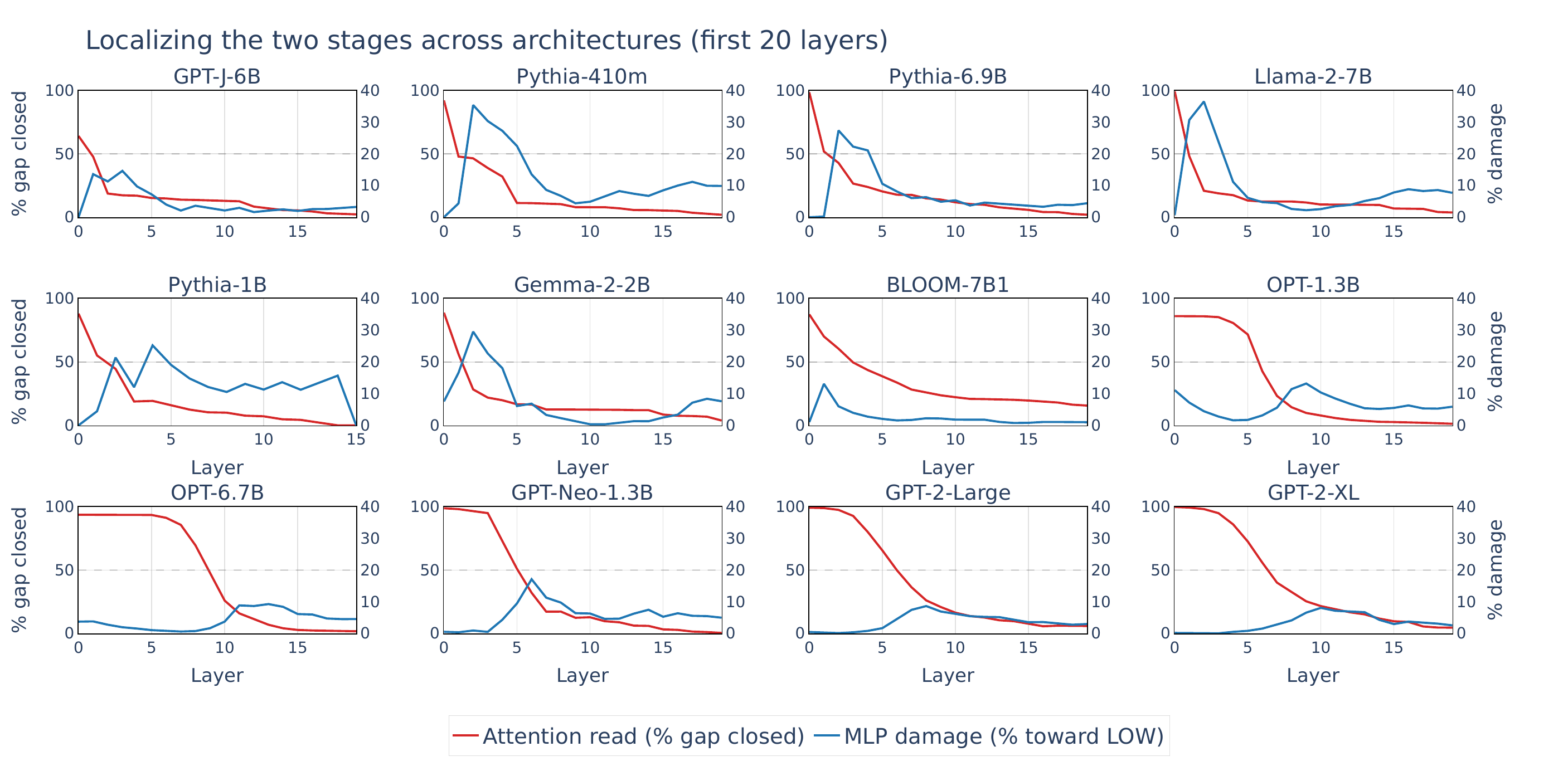}
    \caption{Localizing the two-stage mechanism across twelve
    architectures. 
    Red curve (left y-axis): percentage of the     \metricsrc{} gap closed by patching \texttt{resid\_post} at position~1 from a successful run into a failed one, layer by layer. 
    Blue curve (right y-axis): percentage damage to    \metric{} when patching \texttt{mlp\_out} from a failed run into a successful one. 
    First 20 layers shown for all models.}
    \label{fig:two_stage_mechanism_per_model}
\end{figure*}

\autoref{fig:two_stage_mechanism_per_model} reports the two
interventions of \S\ref{sec:cross_model} layer by layer for each
of the 12 models. 
The red curve isolates the attention stage: patching position~1's residual stream from a successful run into a failed one transfers any cross-position information attention might still read from that position, so the layer at which the patch stops closing the gap marks when attention has finished
its read. 
The blue curve isolates the MLP stage: patching \texttt{mlp\_out} from a failed run into a successful one replaces the MLP's contribution at that layer, so the resulting damage measures how much composition-specific output the MLP is producing there.
The two stages are temporally adjacent in every model: the blue peak sits within or just after the red decay window, never before it. 
Concentrated-regime RoPE models complete both stages within the first $\sim$5 layers, with red decaying by Layer~2--3 and blue peaking at Layer~2--4. 
Distributed-regime models spread both stages across layers 5--12: the red curve decays gradually from a high plateau, and the blue peak shifts proportionally later.
Bloom-7B1 sits between the two regimes.
Its red curve decays gradually over 15+ layers rather than sharply within 2--3, and its blue peak ($\sim$13\% at Layer~1) is smaller than the peaks of the RoPE models. 
The two-stage ordering is intact, but the cross-position read is spread over a much wider window, consistent with ALiBi encoding relative position less sharply than RoPE while still providing relative-position access in attention.

%% file: appendix/gap_closed_per_model.tex
\section{Cross-Architecture Gap-Closed Patching}
\label{app:cross_arch_gap_closed}

\begin{figure*}
    \centering
    \includegraphics[width=1\linewidth]{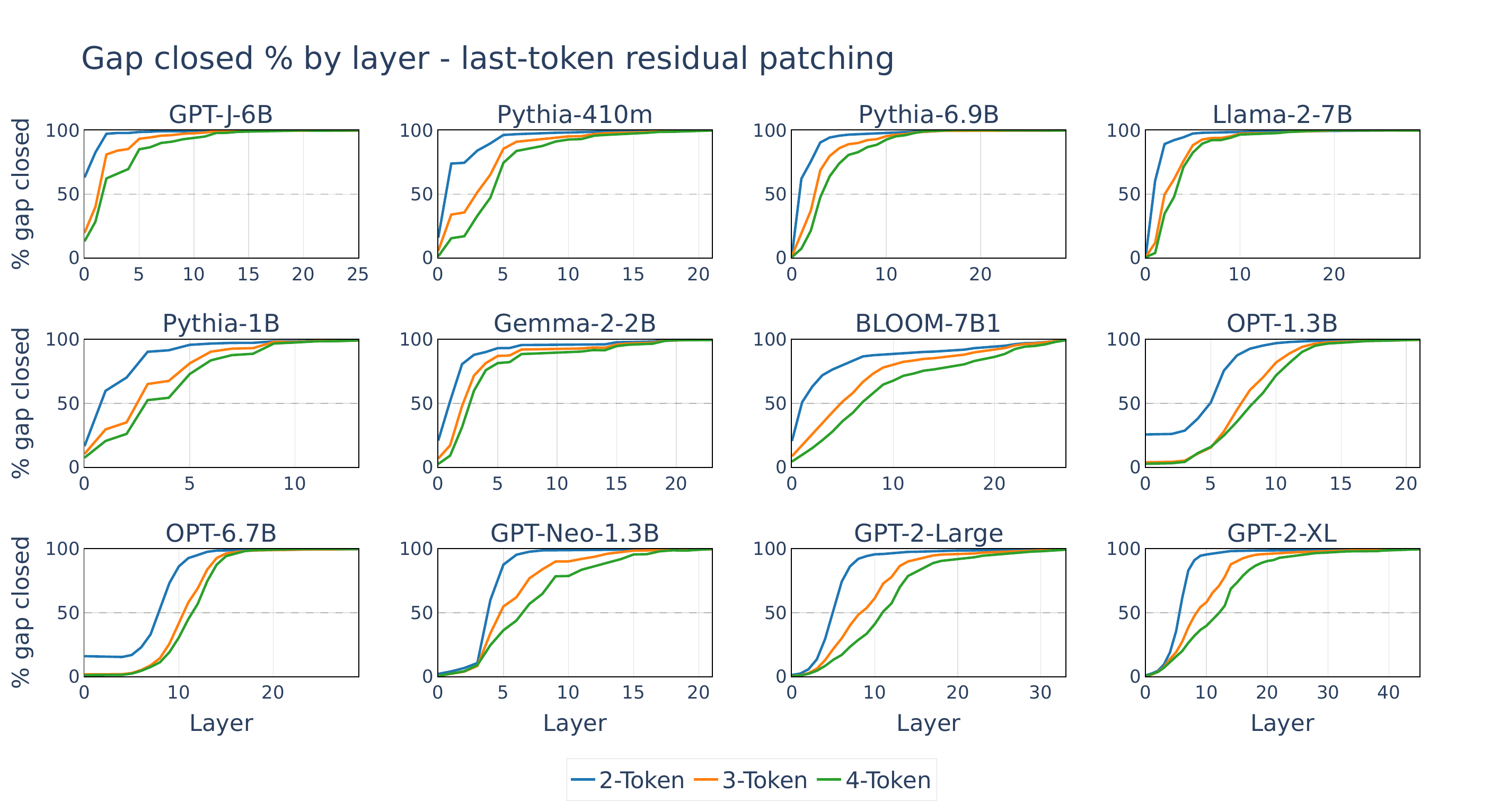}  
    \caption{Percentage of \metricsrc{} gap closed per layer via \texttt{resid\_post} patching at the final subword position, for words of length $k \in \{2, 3, 4\}$ tokens. 
    One panel per model. 
    First-layer indices at which each curve reaches 80\% gap-closed are listed in \autoref{tab:models_depth_scaling_table}.}
    \label{fig:gap_closed_per_model}
\end{figure*}

\begin{table}
\centering
\small
\setlength{\tabcolsep}{2pt}
\begin{tabular}{llccccc}
\toprule
\textbf{Model} & \textbf{PE} & \textbf{Layers} & \multicolumn{3}{c}{$\boldsymbol{\ell^*}$(80\% g-c depth)} \\
\cmidrule(lr){4-6}
 &  &  & $k{=}2$ & $k{=}3$ & $k{=}4$ \\
\midrule
\multicolumn{6}{l}{\textit{Concentrated}} \\
GPT-J-6B      & RoPE  & 28 & 1  & 2  & 5  \\
Pythia-410M   & RoPE  & 24 & 3  & 5  & 6  \\
Pythia-6.9B   & RoPE  & 32 & 3  & 5  & 6  \\
Llama2-7B     & RoPE  & 32 & 2  & 5  & 5  \\
Pythia-1B     & RoPE  & 16 & 3  & 5  & 6  \\
Gemma-2-2B    & RoPE  & 26 & 2  & 4  & 5  \\
\midrule
\multicolumn{6}{l}{\textit{Intermediate (ALiBi)}} \\
Bloom-7B1     & ALiBi & 30 & 5  & 10 & 17 \\
\midrule
\multicolumn{6}{l}{\textit{Distributed}} \\
OPT-1.3B      & Learned & 24 & 7  & 10 & 11 \\
OPT-6.7B      & Learned & 32 & 10 & 13 & 14 \\
GPT-Neo-1.3B  & Learned & 24 & 5  & 8  & 11 \\
GPT2-Large    & Learned & 36 & 7  & 13 & 15 \\
GPT2-XL       & Learned & 48 & 7  & 14 & 17 \\
\bottomrule
\end{tabular}
\caption{Depth required for \detok{} as a function of word length, across all 12 models.
$\ell^*_{80\%}$ is the first layer at which
patching \texttt{resid\_post} at the last subword position closes $\geq 80\%$ of the \metricsrc{} gap.}
\label{tab:models_depth_scaling_table}
\end{table}

Section \ref{sec:cross_model} introduced the concentrated/distributed regime distinction using gap-closed-80\% depth on 2-token words.
This appendix extends the measurement to 3- and 4-token words and reports per-model $\ell^*_{80\%}$ values, which fix the probe depth used in \S\ref{sec:probe}.

\autoref{fig:gap_closed_per_model} shows that the RoPE/learned-PE split holds at every word length. 
The six RoPE models accumulate gap-closed sharply, rising to 80\% within the first $\sim$15\% of network depth and then plateauing. 
Learned-PE models accumulate gradually, rising over roughly the first third of the network before plateauing.
\autoref{tab:models_depth_scaling_table} reports the per-model $\ell^*_{80\%}$ for $k \in \{2, 3, 4\}$. 
Among the RoPE models, $\ell^*_{80\%}$ stays in single digits across all word lengths (5--6 layers at $k{=}4$).
Among the learned-PE models, $\ell^*_{80\%}$ at $k{=}4$ ranges from 11 to 17 layers.
Bloom-7B1 (ALiBi) tracks the concentrated regime at $k{=}2$ ($\ell^*_{80\%}{=}5$) but drifts toward distributed behavior at $k{=}4$ ($\ell^*_{80\%}{=}17$), reaching depths comparable to the slowest learned-PE model (GPT2-XL). 
This is consistent with the BLOOM observation in \S\ref{sec:cross_model}: ALiBi provides relative-position access in attention, but encodes position less sharply than RoPE, and the gap to learned-PE models narrows as the cross-position composition window widens with word length.

%% file: appendix/cross_arch_probe.tex
\section{Cross-Architecture Probe Results}
\label{app:cross_arch_probe}

\begin{figure*}[t]
    \centering
    \includegraphics[width=1\linewidth]{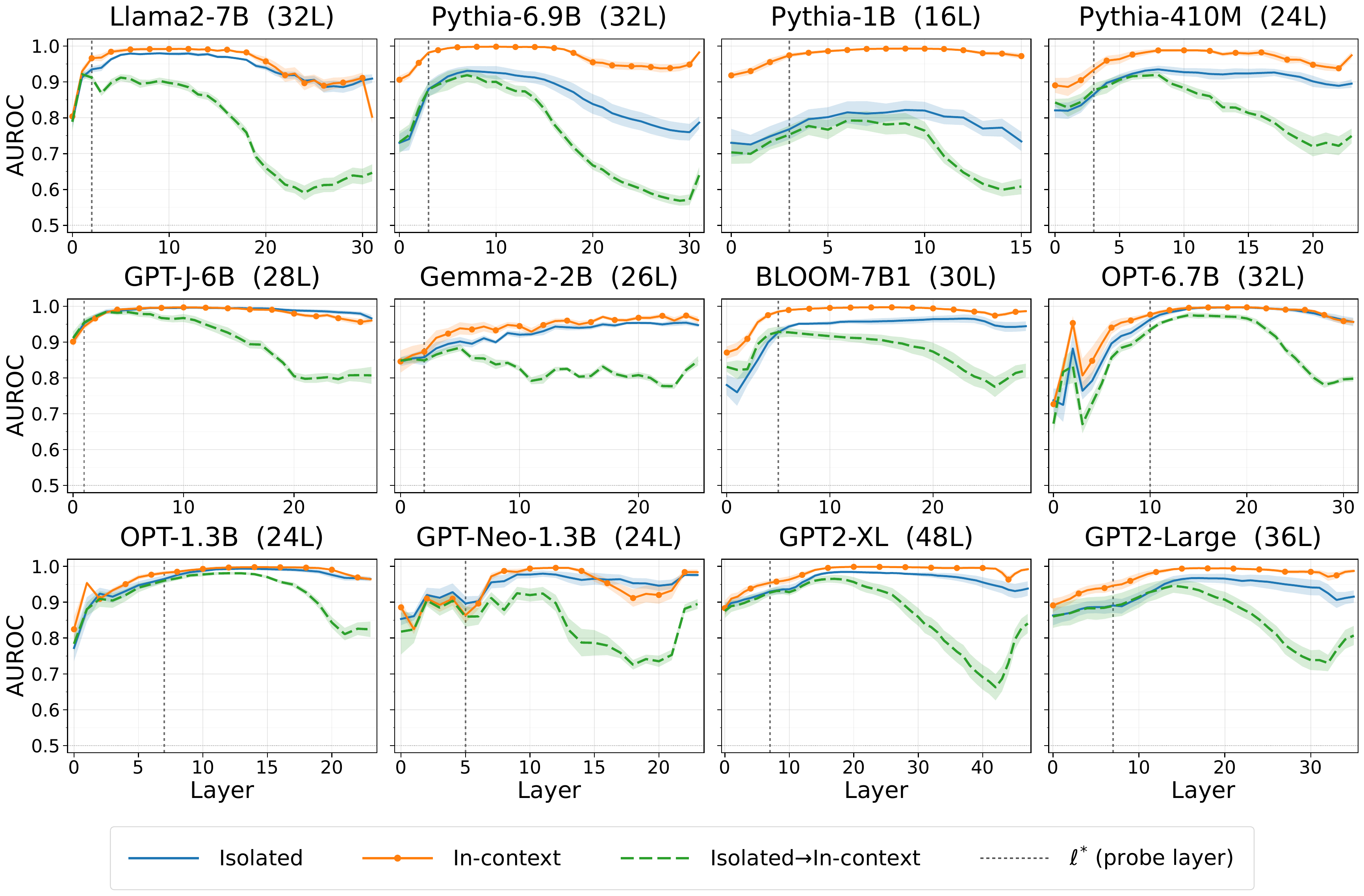}
    \caption{Layerwise probe AUROC for all 12 architectures at
    $k{=}2$. 
    Solid blue: Isolated probe. 
    Solid orange: In-context probe. 
    Dashed green: Isolated$\rightarrow$In-context (probe fit on Isolated activations, evaluated on In-context activations of held-out words). 
    Shaded regions show 95\% confidence intervals across 5 folds. 
    Vertical dotted line: $l^*$ per model.
    The Llama2-7B pattern holds across every architecture: Isolated and In-context rise to a plateau by $l^*$ and stay high;
    Isolated$\rightarrow$In-context tracks them through $l^*$ and decays through middle layers as preceding-token context aggregates into the last subword position.}
    \label{fig:cross_arch_probe}
\end{figure*}

\autoref{fig:cross_arch_probe} reports layerwise AUROC for all 12 architectures at $k{=}2$, replicating the three-curve analysis
of \autoref{fig:llama2_probe_curves}. 
We elaborate here on choices in the probe pipeline and on per-model observations.

\paragraph{Per-model behavior.} 
Peak Isolated AUROC at $l^*$ ranges from $\sim$0.76 (Pythia-1B) to $\sim$0.94 (Llama2-7B, GPT-J-6B, OPT-1.3B). 
Peak In-context AUROC ranges from $\sim$0.86 (Gemma-2-2B) to $\sim$0.99. 
Pythia models show a persistent Isolated/In-context gap, most pronounced in Pythia-1B and Pythia-6.9B (Isolated$\sim$0.75--0.80, In-context$>$0.95). 
Gemma-2-2B is the sole exception where all three curves remain within $\sim$0.05 AUROC of each other across all layers. 
While the Isolated$\rightarrow$In-context decay is universal, its depth varies considerably by model: it is sharpest in Pythia-6.9B, GPT2-XL, and OPT-6.7B (dropping below 0.7), and shallowest in GPT-J-6B and Gemma-2-2B (remaining above 0.8).
This variation suggests that the rate at which context overwrites the early-layer composition signal is architecture-dependent.